\pdfoutput=1

\documentclass[runningheads]{llncs}
\usepackage{microtype}
\usepackage{graphicx}
\usepackage{booktabs} 
\usepackage{multirow}
\usepackage{amsmath}
\usepackage{array}
\usepackage{url}
\usepackage{algorithm,algorithmicx,algpseudocode}
\usepackage{subcaption}
\usepackage{amsfonts}
\newcommand{\E}{\mathbb{E}}

\begin{document}
\title{Quality Evaluation of GANs Using Cross Local Intrinsic Dimensionality}
%
%
\author{Sukarna Barua\inst{1} \and
Xingjun Ma\inst{1} \and
Sarah Monazam Erfani\inst{1} \and
Michael E. Houle\inst{2} \and 
James Bailey\inst{1}}
\authorrunning{S. Barua et al.}
%
\institute{The University of Melbourne, Victoria 3010, Australia \and
National Institute of Informatics, Tokyo 101-8430, Japan
}
\maketitle              
\begin{abstract}
Generative Adversarial Networks (GANs) are an elegant mechanism for data generation.  However, a key challenge when using GANs is how to best measure their ability to generate realistic data. In this paper, we demonstrate that an intrinsic dimensional characterization of the data space learned by a GAN model leads to an effective evaluation metric for GAN quality.  
In particular, we propose a new evaluation measure, CrossLID, that assesses the local intrinsic dimensionality (LID) of real-world data with respect to neighborhoods found in GAN-generated samples.  
Intuitively, CrossLID measures the degree to which manifolds of two data distributions coincide with each other.
In experiments on 4 benchmark image datasets, we compare our proposed measure to several state-of-the-art evaluation metrics. Our experiments show that CrossLID is strongly correlated with the progress of GAN training, is sensitive to mode collapse, is robust to small-scale noise and image transformations, and robust to sample size. Furthermore, we show how CrossLID can be used within the GAN training process to improve generation quality.
\end{abstract}
\makeatletter{\renewcommand*{\@makefnmark}{}
\footnotetext{The first and original version of this paper was submitted to ICLR 2019 conference. 
Submission link: \url{https://openreview.net/pdf?id=BJgYl205tQ}}
\makeatother}
\section{Introduction}
Generative Adversarial Networks (GANs) are powerful models for data generation, composed of two neural networks, known as the \textit{generator} and the \textit{discriminator}. The generator maps random noise vectors to locations in the data domain in an attempt to approximate the distribution of the real-world (or real) data. The discriminator accepts a data sample and returns a decision as to whether or not the sample is from the real data distribution or was artificially generated. While the discriminator is trained to distinguish real samples from generated ones, the generator's objective is to deceive the discriminator by producing data that cannot be distinguished from real data.  The two networks are jointly trained to optimize an objective function resembling a two-player minimax game.  

GANs were first formulated by~\cite{gan}, and have been applied to tasks such as image generation \cite{largescalegan,laplaciangan,sngan,dcgan} and image inpainting \cite{imageinpainting}.
Despite their elegant theoretical formulation~\cite{gan}, training of GANs can be difficult in practice due to instability issues, such as vanishing gradients and mode collapse~\cite{wgan,gan}. The vanishing gradient problem occurs whenever gradients become too small to allow sufficient progress towards an optimization goal within the allotted number of training iterations. The latter occurs when the generator produces samples for only a limited number of data modes, without covering the full distribution of the real data. 

Deployment of GANs is further complicated by the difficulty of evaluating the quality of their output. Researchers often rely on visual inspection of  generated samples, which is both time-consuming and subjective.  A  quantitative quality metric is clearly desirable, and several such methods do exist \cite{modereggan,gan,fid,proggan,c2st,areganequal,acgan,ganimproved,howgoodismygan}. However, past research has identified various limitations of some existing metrics~\cite{inceptionnote,noteevalgenmodel}, and effective evaluation of GAN models is still an open issue.


In this paper, we show how the data distribution learned by a GAN model can be evaluated in terms of the distributional characteristics within neighborhoods of data samples. With respect to a given location $q$ in the data domain, the {\em Local Intrinsic Dimensionality} (LID) model~\cite{lidhoule} characterizes the order of magnitude of the growth of probability measure with respect to a neighborhood of increasing radius. LID can be regarded as a measure of the discriminability of the distribution of distances to $q$ induced by the global distribution; equivalently, it reveals the intrinsic dimensionality of the local data submanifold tangent to $q$. 


Here, we further generalize LID to a new measure, CrossLID, that assesses the average LID estimate over data samples $q$ from one distribution, with respect to a set of samples from a second distribution. If the two distributions are in perfect alignment, the CrossLID measure would yield an estimate of the average LID value with respect to the common distribution. If the distributions were then progressively separated (such as would happen if their underlying manifolds were moved out of alignment), the CrossLID estimate would tend to increase, and become higher than the average LID estimates of samples $q$ with respect to the distribution from which they were drawn. We show that by applying CrossLID to samples from GAN generated data and real data, we can assess the degree to which the GAN generated data distribution conforms to the real distribution.

\begin{figure}[!ht]
\vspace{-0.1 cm}
\centering
\begin{subfigure}{.23\linewidth}
  \centering  
  \includegraphics[width=\textwidth]{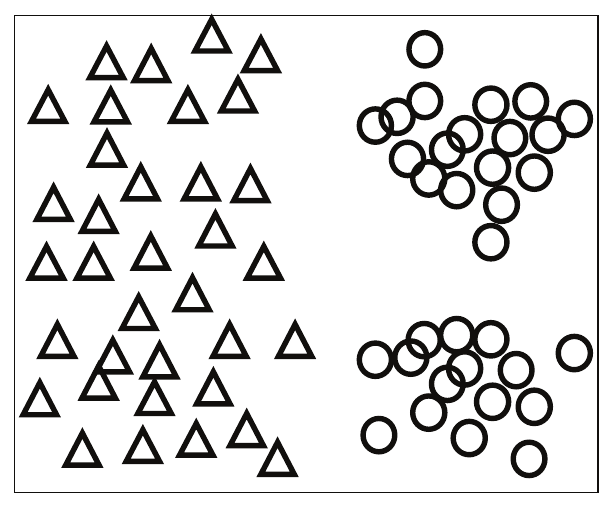}
  \caption{CrossLID=15.16}
\end{subfigure}
\begin{subfigure}{.23\linewidth}
  \centering
  \includegraphics[width=\textwidth]{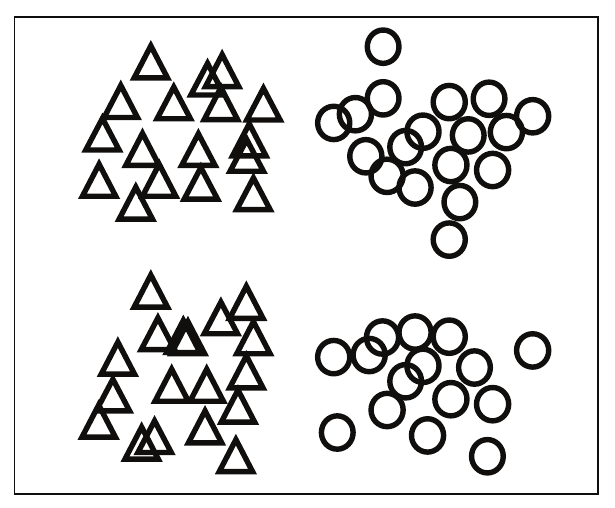}
  \caption{CrossLID=7.33}
\end{subfigure}
\begin{subfigure}{.23\linewidth}
  \centering
  \includegraphics[width=\textwidth]{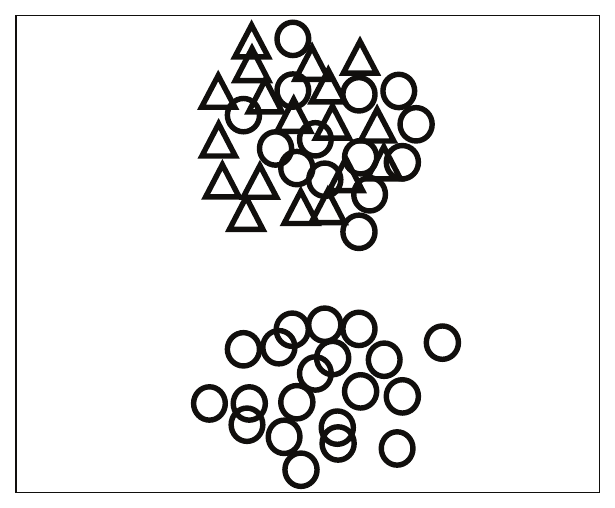}
  \caption{CrossLID=4.78}
\end{subfigure}
\begin{subfigure}{.23\linewidth}
  \centering
  \includegraphics[width=\textwidth]{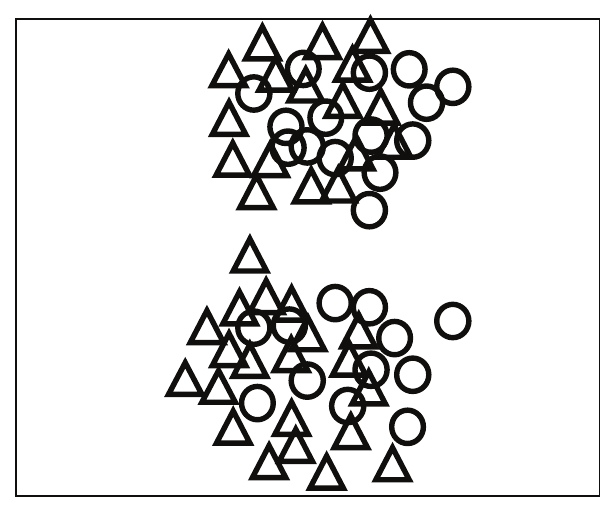}
  \caption{CrossLID=2.10}
\end{subfigure}
\vspace{-0.1 in}
\caption{Four 2D examples showing how GAN-generated data samples (triangles) could relate to a bimodal Gaussian-distributed data set (circles), together with CrossLID scores: (a) generated data distributed uniformly, spatially far from the real data; (b) generated data with two modes, spatially far from the real data; (c) generated data associated with only one mode of the real data; and (d) generated data associated with both modes of the real data (the desired situation).}
\label{fig:lidstudy1}
\vspace{-0.1 in}
\end{figure}

As an illustration of the possible relationships between two distributions, Fig.~\ref{fig:lidstudy1} shows four examples of how a GAN model could learn a bimodal Gaussian distribution. Decreasing CrossLID scores indicate an increasing conformity between the two-mode real data distribution and the generated data distribution. 


The main contributions of the paper are as follows:
\begin{itemize}
\item
We propose CrossLID, a cross estimation technique based on LID that is capable of assessing the alignment of the data embedding learned by the GAN generator with that of the real data distribution. 
\item
We show how CrossLID can be used to avoid mode collapse during GAN training, to identify classes for which some modes are not well-covered by learning.  We also show how this knowledge can then be used to bias the GAN discriminator via an oversampling strategy to improve its performance on such classes. 
\item
Experimentation showing that our CrossLID measure is well correlated with GAN training progress, and comparison to two state-of-the-art GAN evaluation measures.
%
%
\end{itemize}

\section{Evaluation Metrics for GAN Models}
GAN-based learning is an extensively researched area.  Here, we briefly review the topic most relevant to our work, evaluation metrics for GAN models. Past research has employed several different metrics, including log-likelihood measures~\cite{gan}, the Inception score (IS)~\cite{ganimproved}, the MODE score~\cite{modereggan}, Kernel MMD~\cite{kmmd}, the MS-SSIM index~\cite{acgan}, the Fr\'echet Inception Distance (FID)~\cite{fid}, the sliced Wasserstein distance~\cite{proggan}, and Classifier Two-Sample Tests~\cite{c2st}.
In our study, we focus on the two most widely used metrics for image data, IS  and  FID, as well as a recently proposed measure, the Geometry Score (GS) \cite{gscore}.

IS uses an associated Inception classifier~\cite{inceptionv3} to extract output class probabilities for each image, and then computes the Kullback-Leibler (KL) divergence of these probabilities with respect to the marginal probabilities of all classes:
\begin{equation}
\small
\text{IS} = \textrm{exp}( \mathbb{E}_{x \sim p_g} D_{KL}( p(y|x) || p(y) ),
\label{eq:inceptionscore}
\end{equation}
here, $x \sim p_G$ implies a sample $x$ drawn from the generator outputs, $p(y|x)$ is the probability distribution over different classes as assigned to sample $x$ by the Inception classifier, $p(y) = \int_{x}(p(y|x)dx$ is the marginal class distribution, and $D_{\textrm{KL}}$ is the KL divergence. IS measures two aspects of a generative model: (1) the images generated should be both clear and highly distinguishable by the classifier, as indicated by low entropy of $p(y|x)$, and 
(2) all classes should have good representation over the set of generated images, which can be indicated by high entropy of $p(y|x)$ when marginalized over $x$. 
A recent study has shown that IS is susceptible to variations in the Inception network weights when trained on different platforms~\cite{inceptionnote}.

FID passes both real and generated images to an Inception classifier, and extracts activations from an intermediate pooling layer. Activations are assumed to follow multidimensional Gaussians parameterized by their means and covariances. The FID is defined as:
\begin{equation}
\small
\text{FID} = (||\mu_I - \mu_G||)_2^2 + \textrm{Tr}(\Sigma_I + \Sigma_G - 2((\Sigma_I\Sigma_G)^{\frac{1}{2}})), 
\label{eq:fidcore}
\end{equation}
where $(\mu_I,\Sigma_I)$ and $(\mu_G,\Sigma_G)$ represent the mean and covariance of activations for real and generated data samples, respectively. Compared to IS, FID has been shown to be more sensitive to mode collapse and noise in the outputs; however, it also requires an external Inception classifier for its calculation \cite{fid}.

 GS 
 assesses the conformity between manifolds of real and generated data, in terms of the persistence of certain topological properties in a manifold approximation process. The topological relationships are extracted in terms of the counts of 1-dimensional loops in a graph structure built up from proximity relationships as a distance threshold is increased. Although it may be indirectly sensitive to variations in the dimensionality of the manifolds, GS explicitly rewards only matches in terms of the specific topology of these loop structures in approximations of the manifolds. However, due to its strictly topological nature, GS is insensitive to differences in relative embedding distances or orientations within the manifold. This issue is acknowledged by the authors, who advise that GS would be best suited for use in conjunction with other metrics~\cite{gscore}.

\section{Local Intrinsic Dimensionality (LID)}
LID is an expansion-based measure of intrinsic dimensionality within the vicinity of some reference point $q$~\cite{lidhoule}. Intuitively, in Euclidean space, the volume of a $D$-dimensional ball grows proportionally to $r^D$ when its size is scaled by a factor of $r$. From the above rate of volume growth with radius, the dimension $D$ can be deduced from two volume measurements as: $V_2/V_1 = (r_2/r_1)^D \Rightarrow D = \ln(V_2/V_1)/\ln(r_2/r_1)$. Transferring this concept to smooth functions leads to the formal definition of LID.

\textbf{Definition of LID:}
Let $F$ be a positive and continuously differentiable function over some open interval containing $r>0$. The LID of $F$ at $r$ is defined as:
\begin{equation}
\small
\begin{split}
\text{LID}_F(r)
:=
r\frac{F'(r)}{F(r)} &
=
\lim_{\epsilon \to 0^{+}} 
\frac{\ln\,(F((1+\epsilon)r)/F(r))}{\ln\,(1+\epsilon)}
=
\lim_{\epsilon \to 0^{+}}
\frac{F((1+\epsilon)r)-F(r)}{\epsilon\,F(r)}
,
\end{split}
\label{eq:lidhoule}
\end{equation}
wherever the limits exists. The local intrinsic dimensionality of $F$ is then: 
\begin{equation}
\small
\text{LID}^{*}_F=\lim_{r \to 0^{+}} \text{LID}_F(r).
\end{equation}
In our context, and as originally proposed in~\cite{lidhoule}, we are interested in functions that are the distributions of distances induced by some global distribution of data points: for each data sample generated with respect to the global distribution, its distance to a predetermined reference point $q$ determines a sample from the local distance distribution associated with $q$.

The LID model has the interesting property that the definition can be motivated in two different ways.
The first limit stated in the definition follows from a modeling of the growth of probability measure in a small expanding neighborhood of the origin $q$: as the radius $r$ increases, the amount of data encountered can be expected to grow proportionally to the $r$-th power of the intrinsic dimension. Although the LID model is oblivious of the representational dimension of the data domain, in the setting of a uniform distribution with a local manifold of dimension $m$, if $F$ is the distribution of distances to a reference point in the relative interior of the manifold, then $\text{LID}^{*}_F = m$. For more information on the formal definition and properties of LID see~\cite{lidhoule1,lidhoule2}.

The second limit expresses the (in)discriminability of $F$ when interpreted as a distance measure evaluated at distance $r$ (with low values of $\text{LID}_F(r)$ indicating higher discriminability). As implied by Eq.~\ref{eq:lidhoule}, the LID framework is extremely convenient in that the local intrinsic dimensionality and the discriminability of distance measures are shown to be equivalent and interchangeable concepts. 

\textbf{Estimating LID:} 
LID is a generalization of pre-existing expansion-based measures which implicitly use neighborhood set sizes as a proxy for probability measure. These earlier models include the expansion dimension \cite{ed} and its variants \cite{ged}, and the minimum neighbor distance (MinD)~\cite{mind}, all of which have been shown to be crude estimators of LID~\cite{lidestj}. 
Although the popular estimator due to \cite{lidestlevina} can be regarded as a smoothed version of LID, its derivation depends on the assumption that the observed data can be treated as a homogeneous Poisson process. However, the only assumptions made by the LID model is that the underlying (distribution) function be continuously differentiable.
For this work, we use the Maximum Likelihood Estimator (MLE) of LID as proposed in \cite{lidestj}, due to its ease of implementation and its superior convergence properties relative to the other estimators studied there. 

Given a set of data points $X$, and a distinguished data sample $x$, the MLE estimator of LID is:
\begin{equation}
\small
\begin{split}
\text{LID}\,(x; X) &
=
-\Big (\frac{1}{k} \sum_{i=1}^{k} \ln\frac{r_i(x; X)}{r_{max}(x; X)} \Big )^{-1}
= 
\Big (\ln r_{max}(x; X) - \frac{1}{k} \sum_{i=1}^{k} \ln r_i(x; X) \Big )^{-1}
\,
,
\end{split}
\label{eq:lidmle}
\end{equation}
where $k$ is the neighborhood size, $r_i(x; X)$ is the distance from $x$ to its $i$-th nearest neighbor in $X\setminus\{x\}$, and $r_{\textrm{max}}(x; X)$ denotes the maximum distance within the neighborhood (which by convention can be $r_k(x;X)$). 
Due to the deep equivalence between the LID model and the statistical theory of extreme values (EVT) shown in~\cite{lidhoule1,lidestj}, the first of the two equivalent formulations in Eq.~\ref{eq:lidmle} coincides with the well-known Hill estimator of scale derived from EVT~\cite{Hill75}. As can be seen from the second formulation, the reciprocal of the MLE estimator assesses the discriminability within the $k$-NN set of $x$ as the difference between the maximum and mean of log-distance values.
Note that in these estimators, no explicit knowledge of the underlying function $F$ is needed - this information is implicit in the distribution of neighbor distances themselves.



LID can characterize the intrinsic dimensionality of the  data submanifold in the vicinity of a distinguished point $x$. The ${\text{LID}}(x;X)$ values of all data samples $x$ from a dataset $X$ can thus be averaged to characterize the overall intrinsic dimensionality of the manifold within which $X$ resides.   
In \cite{lidhoule2,romano2016measuring}, it was shown that this type of average is in fact an estimator of the correlation dimension over the sample domain (or manifold).
Henceforth, whenever the context set $X$ is understood, we will use the simplified notation ${\text{LID}}(x)$ to refer to ${\text{LID}}(x;X)$, and to denote the average of these estimates over all $x\in X$ by ${\text{LID}}(X)$.

\section{Evaluating GANs via Cross Local Intrinsic Dimensionality}
\label{sec:evaluating_gans_crosslid}

We propose a new measure, CrossLID, that evaluates the conformity between
a real distribution $p_{R}$ and a GAN-generated distribution $p_{G}$, as derived from the profiles of distances from samples of one distribution to samples of the other distribution.
As illustrated in Fig.~\ref{fig:lidstudy1}, our intuition is that if two distributions are similar, then the distance profiles of a generated sample with respect to a neighborhood of real samples should conform with the profiles observed within the real samples, and vice versa.

\subsection{CrossLID for GAN Model Evaluation}
\label{sec:crosslid}
We generalize the single data distribution based LID metric defined in Eq.~\ref{eq:lidhoule} to a new metric that measures the cross LID characteristics between two distributions. 
Given two sets of samples $A$ and $B$, 
the CrossLID of samples in $A$ with respect to $B$ is defined as:
\begin{equation}
\small
\text{CrossLID}(A; B) = \E_{x \in A} {\text{LID}}(x; B).
\label{eq:crosslid}
\end{equation}
Note that 
$\text{CrossLID}(A; B)$ does not necessarily equal $\text{CrossLID}(B; A)$.

Low $\textrm{CrossLID}(A; B)$ scores indicate low average spatial distance between the elements of $A$ and their neighbor sets in $B$. 
From the second formulation of the LID estimator  in Eq.~\ref{eq:lidmle}, we see that increasing the separation between $A$ and $B$ would result in a reduction in the discriminability of distances between them, as assessed by the difference between the maximum and mean of the log-distances from points of $A$ to their nearest neighbors in $B$ --- thereby increasing the CrossLID score. As a simplified example, consider the case where a positive correction $d$ is added to each of the distances from some reference sample $x\in A$ with respect to its neighbors in $B$. This distance correction would cause the reciprocal of the LID estimate defined in Eq.~\ref{eq:lidmle} to become
$\ln(r_{\textrm{max}}{+}d)-\frac{1}{k} \sum_{i=1}^{k} \ln(r_i{+}d)$, 
which leads to an increase in the estimate of $\text{LID}$ 
when $d>0$, and a decrease when $d<0$.
Thus, a good alignment between $A$ and $B$ is revealed by good discriminability (low LID) of the distance distributions induced by one set ($B$) relative to the members of the other ($A$).
In general, CrossLID differs from LID in its sensitivity to differences in spatial position  and orientation of the respective manifolds within which $A$ and $B$ reside (see Suppl. Sec. \ref{sec:manifoldposorientationcrosslid}).


Low values of $\textrm{CrossLID}(A; B)$ also indicate good coverage of the domain of $A$ by elements of $B$. To see why, consider what would happen if this were not the case: if the samples in $B$ did not provide good coverage of all modes of the underlying distribution of $A$, there would be a significant number of samples in $A$ whose distances to its nearest neighbors in $B$ would be excessively large in comparison to an alternative set $B'$ providing better coverage of $A$ (see Fig.~\ref{fig:lidstudy1}c and \ref{fig:lidstudy1}d for an example). As discussed above, this increase in the distance profile would likely lead to an increase in many of the individual LID estimates that contribute to the CrossLID score.

Given a set of samples $X_R$ from a real data distribution, and a set of samples $X_G$ from the GAN-generated distribution, a low value of $\text{CrossLID}(X_{R}; X_{G})$ indicates a good alignment between the manifold associated with $X_G$ and the manifold associated with $X_R$, as well as an avoidance of mode collapse in the generation of $X_G$. 
It should be noted, however, that $\text{CrossLID}(X_{G}; X_{R})$ (in contrast to $\text{CrossLID}(X_{R}; X_{G})$) does not indicate good coverage, and thus is not sensitive to mode collapse.
Since low values of $\text{CrossLID}(X_{R}; X_{G})$ encourage a good integration of generated data into the submanifolds with respect to these learned representations, and an avoidance of mode collapse in sample generation,
$\text{CrossLID}(X_{R}; X_{G})$ is a good candidate measure for evaluating GAN learning processes. 



As CrossLID is local measure rather than global, it also allows targeted quality assessment of GANs for refined sample groups (a subset of real data) of interest. For example, for a specific mode ($X_{R}^{m}$) from the real samples based on either cluster or class information, $\text{CrossLID}(X_{R}^{m}; X_{G})$ can be used to assess how well the GAN model learns the submanifold of this particular mode. CrossLID can therefore be exploited to detect and mitigate underlearned modes in GAN training (explored further in Sec.~\ref{sec:lidoptapproach}).

\subsection{Effective Estimation of CrossLID}
\label{sec:cross_estimation}
We next discuss 2 important aspects in CrossLID estimation: (1) the choice of feature space where CrossLID is computed, and (2) the choice of appropriate sample and neighborhood sizes for accurate and efficient CrossLID estimation.

\textbf{Deep Feature Space for CrossLID Estimation:}
The representations that define the underlying manifold of a data distribution are well learned in the deep representation space. Recent work in representation learning \cite{goodfellow2016deep}, adversarial detection \cite{lidadversarial} and noisy label learning \cite{ddl} has shown that DNNs can effectively map high-dimensional inputs to low-dimensional submanifolds at different intermediate layers of the network. We denote the output of such a layer as a function $f(x)$, and estimate CrossLID in the deep feature space as:
\begin{equation}
\small
\begin{split}
\text{CrossLID}(f(X_{R});f(X_{G})) 
 = \frac{1}{|X_{R}|}\sum_{x \in X_{R}}
\Big (\ln r_{\textrm{max}}(f(x),f(X_{G}))
\mbox{~~~~~~~~~~~~} \\
- \frac{1}{k} \sum_{i=1}^{k} \ln r_i(f(x),f(X_{G})) \Big )^{-1}
\,
.
\end{split}
\label{eq:crosslidmlesetfs}
\end{equation}

It should be noted that successful learning by the GAN discriminator would entail the learning of a mapping $f$ for which the intrinsic dimensionality of $f(X_R)$ is relatively low, and the local discriminability is relatively high. This encourages the GAN generator to produce samples for which $\text{CrossLID}(f(X_{R});f(X_{G}))$ is also low,  further enhancing the value of CrossLID in GAN evaluation and training. 

The transformation $f(x)$ can be computed by training an external network separately on the real data distribution, such as the Inception network used by IS and FID, and then extracting feature vectors from an intermediate layer of the network. In Sec. \ref{sec:experiments_crosslid} we will show how such feature extractors work well for the estimation of CrossLID. Note that CrossLID can be computed using a single forward pass of the feature extractor network --- no backward pass is needed.

\textbf{Sample Size and Neighborhood for CrossLID Estimation:}
Searching for the $k$-nearest neighbors of all samples of $X_{R}$ within the entire GAN-generated dataset $X_{G}$ can be prohibitively expensive. Recent works using the  LID measure in adversarial detection \cite{lidadversarial} and noisy label learning \cite{ddl} have demonstrated that LID estimation at the deep feature level can be effectively performed within small batches of training samples --- with neighborhood sizes as small as $k=20$ drawn from batches of 100 samples. For the estimation of $\text{CrossLID}(f(X_{R});f(X_{G}))$, we use $|X_R|=20000$ samples from the real training dataset, and $|X_{G}|=20000$ GAN-generated samples. To reduce computational complexity, we search $k=100$ nearest neighbors of each $f(x)$, where $x\in X_R$, within a batch of 1000 samples randomly chosen from $f(X_{G})$, and use the distances from $f(x)$ to these $k=100$ nearest neighbors to estimate $\text{CrossLID}(f(x);f(X_{G}))$. The mean of the CrossLID estimates over all 20000 real samples determines the final overall estimate. A larger $k$ tends to result in a higher value of  CrossLID, an effect of the expansion of locality (more details in Suppl. Sec.~\ref{sec:crosslidvaryk}).

\section{Oversampling in GAN training with Mode-wise CrossLID}
\label{sec:lidoptapproach}


A GAN distribution may not equally capture the
distributions of all modes presenting in a real data distribution. 
Due to the inherent randomness in stochastic learning, the decision boundary of the discriminator may be closer to regions of some modes than others at different stages of the training process. 
The closer modes may develop stronger gradients, in which case the generator would learn these modes better than the others. If imbalances in learning can be detected and addressed during training, we could expect a better convergence to good solutions. To achieve this, we propose a GAN training strategy with oversampling based on mode-wise CrossLID scores (as defined in Sec. \ref{sec:crosslid}).

We describe our training strategy in the context of labeled data, where we simply take the classes to be the modes (or clusters if data is unlabeled). 
We compute the average CrossLID score for real samples (w.r.t. generated samples) from each class, and use it to assess how well a class has been learned by the generator --- the lower the CrossLID score, the more effective the learning. To  generate good gradients for all classes during the training, we dynamically modify the input samples of the discriminator by oversampling the poorly learned classes (those with high class-wise CrossLID scores) from the real data distribution. The objective is to bias the discriminator's decision boundary towards the regions of poorly learned classes and to produce stronger gradients for the generator in favor of underlearned classes.

The steps are described in Alg.~\ref{algorithm:crosslid}. From each class $c \in \{1, \cdots, C\}$, we select a subset of  samples $X_{c}^{'}$, of size proportional to a deviation factor $\gamma_c = |\text{CrossLID}(X_{R}^{c}; X_{R}^{c}) - \text{CrossLID}(X_{R}^{c}; X_{G})|/\text{CrossLID}(X_{R}^{c}; X_{R}^{c})$, and augment the original real dataset with the members of $X_{c}^{'}$ for subsequent training. 
$\gamma_c$ measures the relative deviation of the  $\text{CrossLID}(X_{R}^{c}; X_{G})$ score from the self-CrossLID score $\text{CrossLID}(X_{R}^{c}; X_{R}^{c})$, i.e, the LID of $X_{R}^{c}$. When the GAN  has already fully learned the distribution of a given class (i.e., $\text{CrossLID}(X_{R}^{c}; X_{G}) = \text{CrossLID}(X_{R}^{c}; X_{R}^{c})$), $\gamma_c=0$, indicating that no oversampling will be applied to this class.

\begin{algorithm}[ht!]
\small
\caption{GAN training with mode-wise oversampling}
\label{algorithm:crosslid}
\begin{algorithmic}[1]
    
    \For {every $T$ generator iterations}
    \State Generate $N_1$ GAN samples $X_{G}$. 
    \For {$c$ in $\{1, \cdots, C\}$} \Comment{$C$: number of classes.}
    \State Sample $N_2$ real samples $X_{R}^{c}$ from class $c$ 
    \State $\gamma_c = |\text{CrossLID}(X_{R}^{c}; X_{R}^{c}) - \text{CrossLID}(X_{R}^{c}; X_{G})|/\text{CrossLID}(X_{R}^{c}; X_{R}^{c})$
    \EndFor
    \State $\gamma_c=\gamma_c/\sum_{j=1}^{C}\gamma_j$, for $c \in \{1, \dots, C\}$. \Comment{Normalization} 
    \State $X_{\textrm{aug}} = \{X_1^{'}, \cdots, X_C^{'}\} \cup X_{R}$ where $X_c^{'}$ is a random sample from $X_R^c$ of size 
    \Statex \hspace{1cm} $|X_c^{'}| = m \times \gamma_c$ where $m$ is a size parameter
    \State Continue GAN training with $X_{\textrm{aug}}$ for the next $T$ generator iterations.
    \EndFor
 \end{algorithmic}
 \end{algorithm}

Our proposed strategy can effectively deal with the mode collapse issues encountered in GAN training. When the generator learns a class partially, or not at all, it receives a relatively high CrossLID score for that class. In subsequent iterations, the imbalance in learning will be addressed by our oversampling in favor of these classes (Step 8 in Alg.~\ref{algorithm:crosslid}). 

Note that CrossLID guided training can be used for unconditional GAN training that does not explicitly use label information in the generator. CrossLID does not require the knowledge of class or mode information of the GAN generated samples; it only requires the same information of the target dataset only. Unlike CrossLID, other metrics such as Inception score and FID cannot be used for mode-wise performance estimation as they are inherently global estimates. FID can be estimated class-wise (or mode-wise) provided that we know the label (mode) information of the GAN generated samples, which might be available in a (class) conditional GAN training only. Thus, the proposed mode-wise training is more widely applicable using CrossLID than using other metrics, e.g., FID, in different training settings.

\section{Experimental Results}
\subsection{Evaluation of CrossLID as a GAN Quality Metric}
\label{sec:experiments_crosslid}
We first demonstrate that the CrossLID score is well correlated with the training progress of GAN models. We then discuss four characteristics of the CrossLID metric: (1) sensitivity to mode collapse, (2) robustness to small input noise, (3) robustness to small image transformations, and (4) robustness to sample size used for estimation. We also compare CrossLID score with Geometry score, IS and FID. For evaluation, we used 4 benchmark image datasets: MNIST~\cite{lecun1990handwritten}, CIFAR-10~\cite{krizhevsky2009learning}, SVHN~\cite{netzer2011reading}, and ImageNet~\cite{imagenet}.

For the CrossLID score, we used external CNNs trained on the original training set of real images for feature extraction (more details in Suppl. Sec~\ref{sec:cnnfeatureextractor}). To compute the IS, we followed \cite{ganimproved} using the pretrained Inception network, except in the case of MNIST, for which we pretrained a different CNN model as described in \cite{mnistinception}. FID scores were computed as in~\cite{fid}. 
Our code is available at \url{https://www.dropbox.com/s/bqadqzr5plc6xud/CrossLIDTestCode.zip}.

\begin{figure*}[ht!]
\centering
\vspace{-0.2 in}
\begin{tabular}{c}
    \includegraphics[width=122mm]{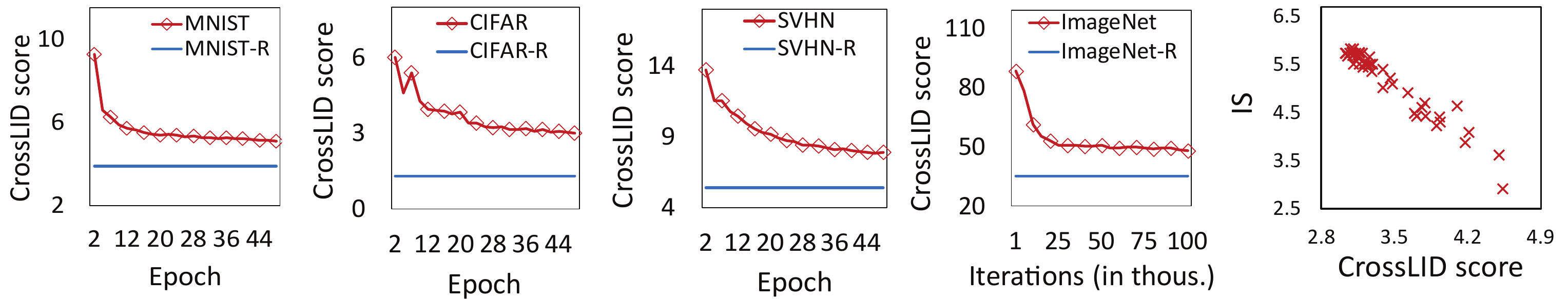} \\
\end{tabular}
\vspace{-0.1 in}
\caption{\emph{Left three}: The CrossLID($X_{R};X_{G}$) score for samples generated by a DCGAN model after different training epochs. Results are shown for the first 50 epochs of training on MNIST, CIFAR-10 and SVHN, and MNIST-R/CIFAR-R/SVHN-R denote the CrossLID($X_{R};X_{R}$) scores for real MNIST/CIFAR-10/SVHN samples. \emph{Fourth from left}: CrossLID score for ImageNet samples generated by a ResNet model (trained using WGAN-GP algorithm) after different generator iterations. \emph{Rightmost}: Correlation between CrossLID and IS for CIFAR-10 dataset, each point is associated with a model at a certain epoch.
}
\label{fig:lidstudy2}
\vspace{-0.2 in}
\end{figure*}

\textbf{Correlation of CrossLID and Training Progress of GANs:}
We show that the CrossLID score is highly correlated with the training progress of GAN models. In the left three subfigures of Fig.~\ref{fig:lidstudy2}, as GAN training proceeds, the CrossLID score decreases (supporting images are reported in Suppl. Sec.~\ref{sec:correlationlidsamplequality} for visual verification of training progress). CrossLID($X_{R};X_{G}$) was estimated over 20,000 generated samples using deep features extracted from the external CNN model. To show that CrossLID metric remains effective for high dimensional datasets, we evaluate it on the $128 \times 128$ pixel ImageNet dataset consisting of 1000 classes, each class having approximately 1300 images. We trained a ResNet model using WGAN-GP~\cite{improvedwgan} algorithm  on the full ImageNet dataset for $100K$ generator iterations and computed CrossLID scores after every 1000 generator iterations. The fourth subfigure from the left in Fig.~\ref{fig:lidstudy2} shows the computed CrossLID scores over different generator iterations confirming that CrossLID score improves (decreases) as GAN training proceeds.

IS is an established metric that was demonstrated to correlate well with human judgment of sample quality~\cite{ganimproved}. The rightmost subfigure in Fig.~\ref{fig:lidstudy2} illustrates the strong negative correlation between CrossLID and IS over different training epochs. We also observed a strong positive correlation of CrossLID with FID (results are reported in Suppl. Sec.~\ref{sec:crosslidfidcorr}). We found that GS does not exhibit a clear correlation with sample quality, which is consistent with its reported insensitivity to differences in embedding distances or orientations \cite{gscore} (see Suppl. Sec.~\ref{sec:geometryscore}). Therefore, we omit  GS from the remainder of the discussion.

\textbf{Sensitivity to Mode Collapse:}
A challenge of GAN training is to overcome mode collapse, which occurs when the generated samples cover only a limited number of modes (not necessarily from the real distribution) instead of learning the entire real data distribution. An effective evaluation metric for GANs should be sensitive to such situations.

We simulate two types of mode collapse by downsampling the training data: (1) \emph{intra-class mode dropping}, which occurs when the GAN generates samples covering all classes, and (2) \emph{inter-class mode dropping}, which occurs when the GAN generates samples from a limited number of classes. 
For both types, we randomly select a subset of $n$ samples from $c$ classes from the original training set (of $N$ samples from $C$ classes), then randomly subsample with replacement from the subset to create a new dataset with the same number of samples $N$ as in the original training set. For the simulation of intra-class mode dropping, we let $c=C$, and vary $n \in [30, 40, 50, 70, 100]$, whereas for inter-class mode dropping we let $n=50$, and vary $c \in [2, 4, 6, 8, 10]$. Overall, for each of the original datasets, we created five new datasets for each type of mode collapse, and computed CrossLID, IS, and FID scores on the new datasets. Note that each of these new datasets has the same number of instances $N$ as the original training set from which it was derived.   
\begin{figure}[ht!]
\centering
\setlength{\tabcolsep}{2pt}
\vspace{-0.2 in}
\begin{tabular}{c}
    \includegraphics[width=90mm]{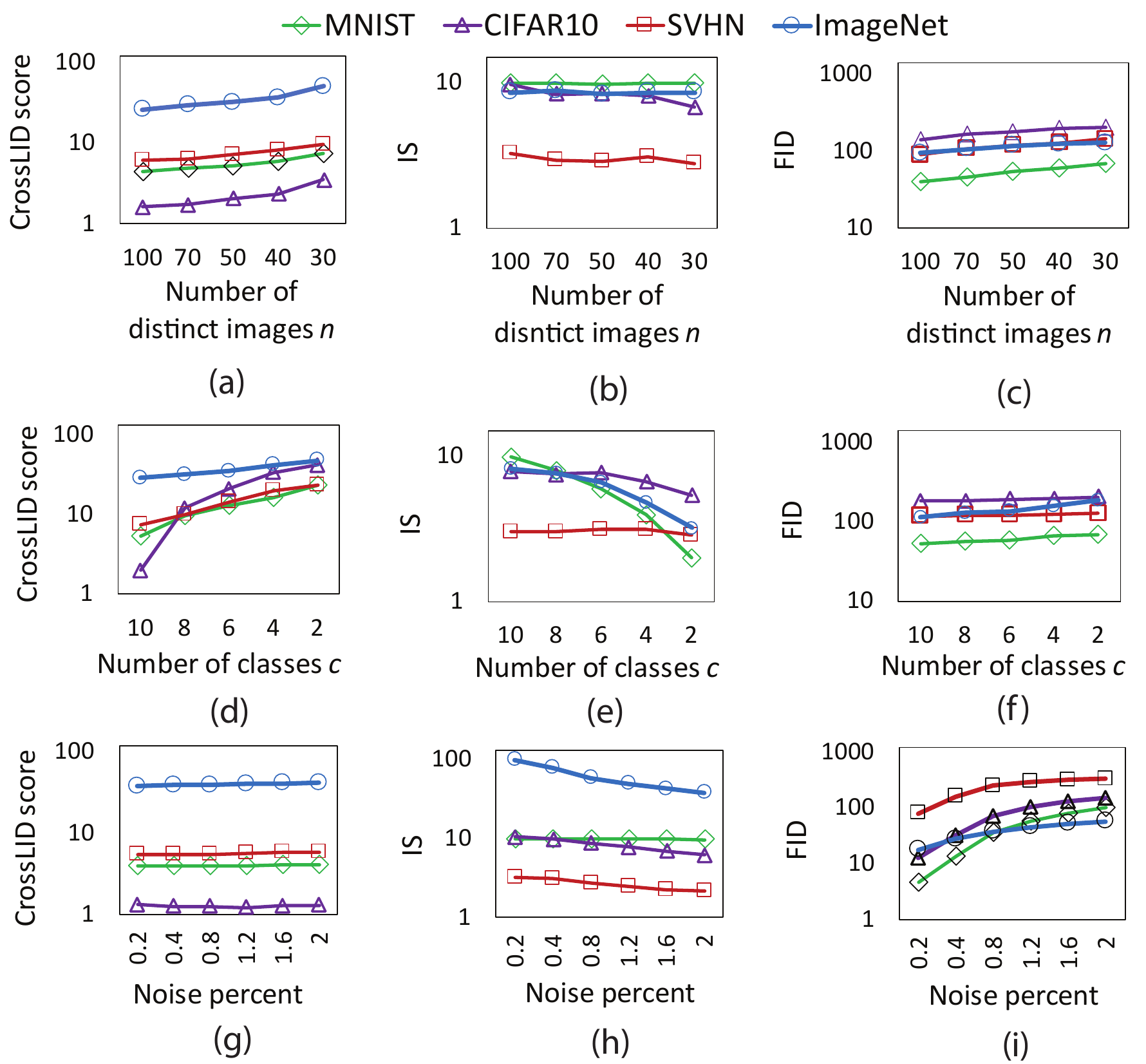} \\
\end{tabular}
\vspace{-0.1 in}
\caption{\emph{(a--c)} Test results for intra-class mode dropping: The CrossLID, IS, and FID scores on varying numbers of unique samples in the datasets. \emph{(d--f)} Test results for inter-class mode dropping. \emph{(g--i)} Robustness to Gaussian noise for CrossLID, IS, and FID. Noise percent indicates the proportion of pixels of GAN images that have been modified with noise.}
\label{fig:lidinceptionfidvarydiversitynoise}
\vspace{-0.2 in}
\end{figure}
As shown in Fig.~\ref{fig:lidinceptionfidvarydiversitynoise}(a), we found that CrossLID is sensitive to different degrees of intra-class mode dropping, but IS failed to identify intra-class mode dropping on MNIST and ImageNet, and responded inconsistently for different levels of intra-class mode dropping on CIFAR-10 and SVHN (Fig.~\ref{fig:lidinceptionfidvarydiversitynoise}(b)). 
FID is also sensitive to intra-class mode dropping (Fig.~\ref{fig:lidinceptionfidvarydiversitynoise}(c)). Similar results were seen for inter-class mode dropping (Fig.~\ref{fig:lidinceptionfidvarydiversitynoise}(d--f)): again, CrossLID was found to be sensitive to increasing levels of inter-class mode dropping, and is more sensitive than FID. Although IS revealed inter-class mode dropping for MNIST and ImageNet, it failed to do so for CIFAR-10 and SVHN. 


\textbf{Robustness to Small Input Noise:}
We examine the robustness of the three metrics to small noise in data which does not alter visual quality greatly. We add noise drawn from a Gaussian distribution with both mean and variance equal to 127.5 (255/2) to a small proportion of pixels in the original images. As shown in Fig.~\ref{fig:lidinceptionfidvarydiversitynoise}(g), CrossLID exhibits small variations as the proportion of modified pixels increases from 0.2\% to 2\%. For example, on CIFAR10 dataset, CrossLID score changes by only 1.2\% at 2\% Guassian noise. In contrast, both IS and FID demonstrate disproportionately large variations, particularly for CIFAR-10, SVHN, and ImageNet (Fig.~\ref{fig:lidinceptionfidvarydiversitynoise}(h) and~\ref{fig:lidinceptionfidvarydiversitynoise}(i)). For example, IS and FID change by 52\% and 48\%, respectively, at only 2\% noise level on CIFAR10. The behavior of the three metrics remain similar even if we normalize the scores with respect to their minimum and maximum values. (Details and further experiments with different noise types are reported in Suppl. Sec.~\ref{sec:robustnessspnoise} and \ref{normalizedscoresnoise}).

\begin{figure*}[ht!]
\centering
\setlength\tabcolsep{2pt}
\vspace{-0.2 in}
\begin{tabular}{c}
    \includegraphics[width=120mm]{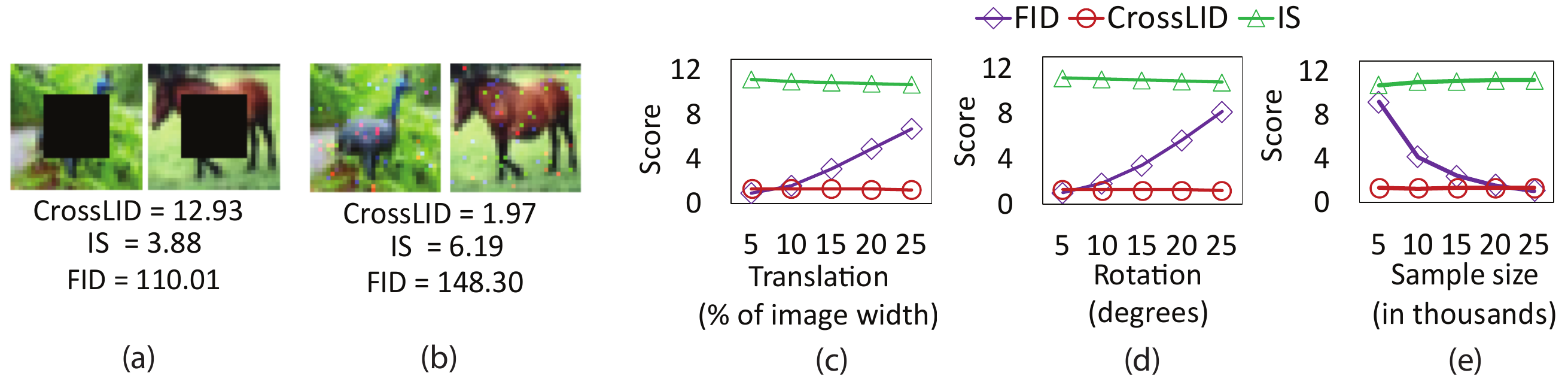} \\
    \vspace{-12.0pt} \\
\end{tabular}
\caption{(a--b) The CrossLID, IS, and FID scores of images with centers occluded by black rectangles (\emph{a}) and with 2\% Gaussian noise (\emph{b}). (c--e) Robustness test of the three metrics on CIFAR-10 dataset to small image transformations including translation, rotation, and sample size used for calculation.}
\vspace{-0.2 in}
\label{fig:lidinceptionfidvarytrans}
\end{figure*}

A potential drawback of high sensitivity to low noise levels is that the metric may respond inconsistently for images with low noise as compared to images of extremely low quality. Consider the figures~\ref{fig:lidinceptionfidvarytrans}(a) and ~\ref{fig:lidinceptionfidvarytrans}(b), wherein we report the three metrics for two specific types of noise: a black rectangle obscuring the center of the images, and 2\% Gaussian noise, respectively. Although the images with Gaussian noise are visually superior to the other ones (with implanted rectangle), by virtue of its lower score, FID rates the obscured images to be of better quality --- quite the opposite to human visual judgment.
In contrast, for this particular scenario, the response of both CrossLID (for which a lower score indicates better quality) and IS (for which a higher score is better) is in line with human assessment. We believe that robustness to small input noise which does not greatly change visual quality of images is a desirable characteristic for a quality measure. Noting that there is as yet no consensus on the issue of whether GAN quality measures should be robust to noise, we pose it as an open problem for the GAN research community to explore. 

\textbf{Robustness to Small Input Transformation:}
We further test the robustness of the metrics to small input transformations. As long as the transformations do not alter the visual appearance of GAN images, a robust metric should be able to give consistent evaluations. This is important in that GAN generated images often exhibit small distortions compared to natural images, and such small imperfections should not significantly detract from the perceived quality of GANs. As demonstrated in figures~\ref{fig:lidinceptionfidvarytrans}(c) and \ref{fig:lidinceptionfidvarytrans}(d), CrossLID and IS conform with each other showing moderate sensitivity to small translations and rotations on CIFAR-10 images. This is reasonable considering that the convolution layers of the feature extractor are expected to learn features which are moderately invariant to small input transformations. However, we find that FID changes drastically with small input transformations (Fig.~\ref{fig:lidinceptionfidvarytrans}(c--d)). FID calculation on different (non-Inception) feature spaces could possibly lead to different behavior; however, this investigation is beyond the scope of this paper.

\textbf{Robustness to Sample Size:}
For the sake of efficiency, it is desirable that GAN quality measures be able to perform well even when computed over relatively small sample sizes. We test the robustness of the three metrics versus sample size on a subset of CIFAR-10 training images. The results are shown in Fig.~\ref{fig:lidinceptionfidvarytrans}(e). CrossLID and IS are moderately stable as the subset size decreases from 25K to 5K, in particular, CrossLID exhibits the least variation. However, the FID score turns out to be quite highly sensitive to the sample size. The lower variation of CrossLID against sample size allows it to be computed on samples of smaller size as compared to what is typically needed by the other two metrics. We note that previous research on FID \cite{areganequal} has noted that it exhibits high variance for low sample sizes, and has hence been recommended only for sufficiently large sample sizes ($>10K$). We have also compared the running times of the three metrics with respect to different sample sizes and found that CrossLID requires the lowest computation time while FID the highest (Details in Suppl. Section~\ref{sec:metricexectime}).

\textbf{Summary:}
Table \ref{tab:lidcompareinceptionfid} summarizes our experimental comparisons of CrossLID, IS, and FID.

\begin{table}[!h]
\small
\setlength\tabcolsep{5pt}
\vspace{-0.4 in}
\caption{Evaluation summary of CrossLID, IS and FID.}
\label{tab:lidcompareinceptionfid}
\begin{center}
\begin{tabular}{m{6.0cm}|ccc}
\hline
EVALUATION CRITERIA & CrossLID & IS & FID\\  \hline \hline
Sensitivity to mode collapse.&High&Low&High\\  \hline
Robustness to small input noise.&High&Low&Low\\ \hline
Robustness to small input transformations. &Moderate&Moderate&Low\\ \hline
Robustness to sample size variation.&High&Moderate&Low \\ \hline
\end{tabular}
\end{center}
\vspace{-0.4 in}
\end{table}


\subsection{Evaluation of the Proposed Oversampling}
\label{sec:agumented_training}
We evaluate the effectiveness of the CrossLID-guided oversampling approach in GAN training. For the MNIST, CIFAR-10 and SVHN datasets, we compare standard versions of the popular DCGAN~\cite{dcgan} and WGAN~\cite{wgan} models to the same models trained with CrossLID-guided oversampling. (Further details of model architecture, experimental settings, and output images can be found in Suppl. Sec. \ref{sec:expsetup}.)

The performances in terms of CrossLID scores are reported in Table~\ref{tab:dcganlidinceptionscores}, where DCGAN+ and WGAN+ refer to training with our proposed oversampling (IS and FID results for these experiments are reported in Suppl. Sec.~\ref{sec:inceptionfidoversamplingexp}). Our training approach achieved comparatively better results than the standard training in terms of CrossLID, IS, and FID, for both DCGANs and WGANs.

\begin{table}[!htb]
\vspace{-0.3 in}
	\small
    \setlength\tabcolsep{2.0pt}
    \begin{center}
    \caption{Performance of oversampling on DCGAN and WGAN.}
    \label{tab:dcganlidinceptionscores}
    \begin{tabular}{c|c|c||c|c}
        \hline
      	&\multicolumn{4}{c}{CrossLID score (lower is better)} \\ \hline
        Dataset&DCGAN&DCGAN+&WGAN&WGAN+ \\ \hline
        MNIST&5.11 $\pm$ 0.02&\textbf{4.96 $\pm$ 0.08}&5.91 $\pm$ 0.02&\textbf{5.26 $\pm$ 0.02} \\ \hline
        CIFAR10&3.00 $\pm$ 0.04&\textbf{2.78 $\pm$ 0.04}&3.70 $\pm$ 0.04&\textbf{3.57 $\pm$ 0.04} \\ \hline
        SVHN&7.40 $\pm$ 0.01&\textbf{7.14 $\pm$ 0.03}&10.14 $\pm$ 0.04&\textbf{9.95 $\pm$ 0.04} \\ \hline
    \end{tabular}
    \end{center}
    \vspace{-0.0 in}
\end{table}

\begin{figure}[!htb]
	\centering
	\vspace{-0.4 in}
	\small
	\setlength\tabcolsep{3pt}
	\begin{tabular}{cccc}
		\includegraphics[width=18mm]{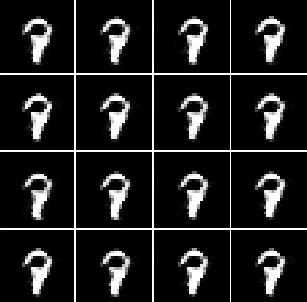} & 
			\includegraphics[width=17.5mm]{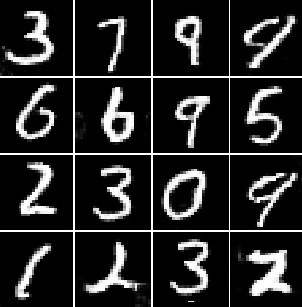} &
	           	\includegraphics[width=18mm]{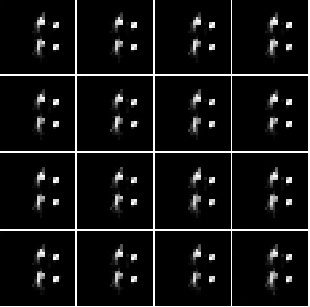} & 
			        \includegraphics[width=18mm]{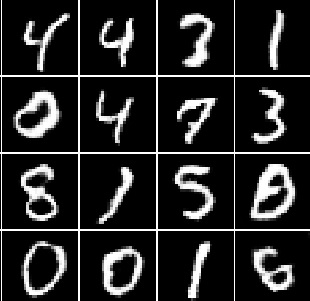} \\ 
		\small{(a) DCGAN} & \small{(b) DCGAN+} & \small{(c) DCGAN} & \small{(d) DCGAN+ }\\
\end{tabular}
\caption{Images generated at the end of the 30-th epoch by DCGAN and DCGAN+ on the MNIST dataset, when batch normalization is removed from the discriminator (a--b) and from both the generator and discriminator (c--d).}
\label{fig:mnistoutputsmodecollapse}
\vspace{-0.2 in}
\end{figure}

\textbf{Effectiveness in Preventing Mode Collapse:} 
As explained in Sec~\ref{sec:lidoptapproach}, our approach can help avoid mode collapse. We next show on MNIST, when the batch-normalization layers were removed from the discriminator, or from both the discriminator and the generator, standard DCGAN training suffered significant mode collapse and failed to learn the full real distribution, as shown in Fig.~\ref{fig:mnistoutputsmodecollapse}(a) and~\ref{fig:mnistoutputsmodecollapse}(c). Our approach, however, was still able to produce high quality images without any sign of mode collapse during the training, as shown in Fig.~\ref{fig:mnistoutputsmodecollapse}(b) and~\ref{fig:mnistoutputsmodecollapse}(d). (Training process and visual inspections reported in Suppl. Sec.~\ref{sec:stability}.)

\section{Conclusion}
We have proposed a new metric for quality evaluation of GANs, based on cross local intrinsic dimensionality (CrossLID). Our measure can effectively assess GAN generation quality and mode collapse in GAN outputs.  It is reasonably robust to input noise, image transformations, and sample size. We also demonstrated a simple oversampling approach based on the mode-wise CrossLID that can improve GAN training and help avoid mode collapse.  

We believe CrossLID is not only a promising new tool for assessing the quality of GANs, but also can help improve GAN training.  We envisage CrossLID can be used as an additional metric for the community to evaluate GAN quality. Unlike IS and FID, CrossLID uses a local perspective rather than global perspective when evaluating sample quality, in that a quality score for each individual GAN generated sample can be computed based on its neighborhood. The advantage of mode-wise performance estimation by CrossLID may be utilized in different GAN models such as conditional and supervised GANs. 


\bigskip

\bibliographystyle{splncs04}

\bibliography{crosslid}

\appendix

\section{The Effect of Manifold Positioning and Orientation on CrossLID}
\label{sec:manifoldposorientationcrosslid}
\begin{figure}[ht!]
\centering
\setlength\tabcolsep{5pt}
\begin{tabular}{cccc}
    \includegraphics[width=0.17\textwidth]{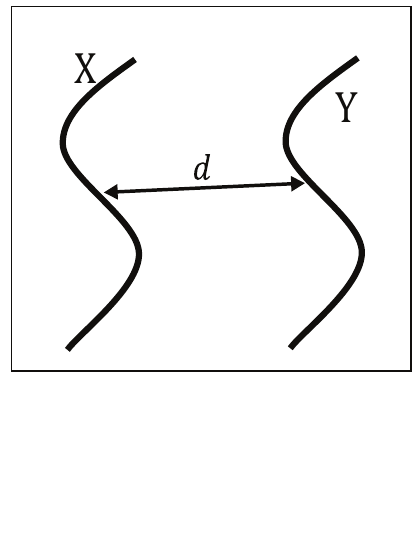}  &
    	 \includegraphics[width=0.25\textwidth]{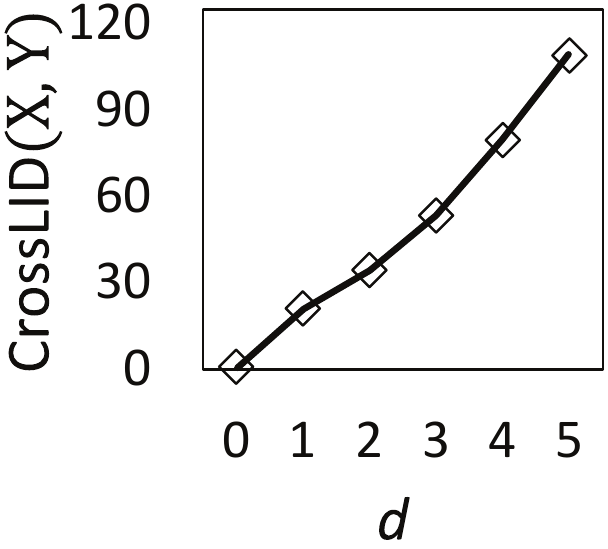} & \includegraphics[width=0.18\textwidth]{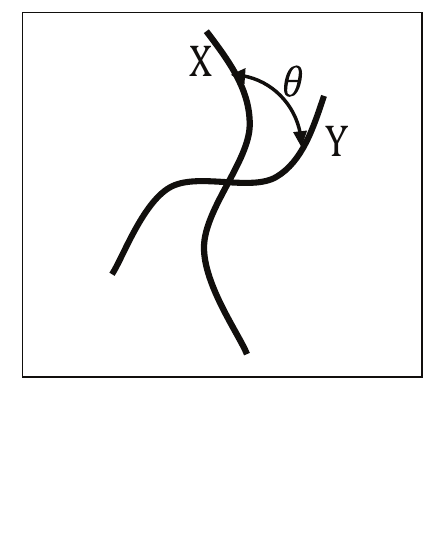} &
        \includegraphics[width=0.25\textwidth]{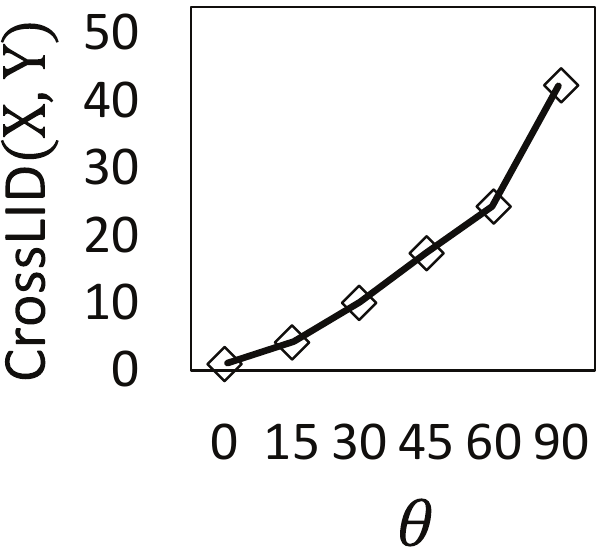}\\
    (a) & (b) & (c) & (d)  \\
\end{tabular}
\vspace{-0.1in}
\caption{(a) Two manifolds $X$ and $Y$ have identical geometric structures, but different positions in space; (b) CrossLID score decreases as $X$ and $Y$ moves closer to each other (decreasing $d$); (c) The same manifolds (as in (a)) but having different orientations (rotated with respect to each other); (d) CrossLID score decreases as the relative orientation angle $\theta$ decreases.}
\label{fig:crosslidmanorien}
\end{figure}

In Figure~\ref{fig:lidstudy1} of Section \ref{sec:evaluating_gans_crosslid} of the main text, we have shown the power of CrossLID in capturing submanifold conformity via a toy example where a GAN model attempts to learn a bimodal Gaussian distribution. Here, we provide more insights into the understanding of CrossLID. Figure~\ref{fig:crosslidmanorien} illustrates how our proposed CrossLID can effectively characterize the conformity of two manifolds ($X$ and $Y$) with identical geometric structures but different positioning or orientation in space. As the two manifolds move closer and closer to each other either in position (decreasing distance $d$) or orientation (decreasing angle $\theta$), the CrossLID scores ($\text{CrossLID}(X;Y)$) tend to decrease, and are close to one ($\text{CrossLID}(X;X)$) when the two manifolds are completely overlapping. This substantiates how effectively CrossLID can measure differences in spatial position and orientation of two manifolds, even when they have similar structure.

\begin{figure}[ht!]
\centering
\begin{tabular}{c}
    	\includegraphics[width=100mm]{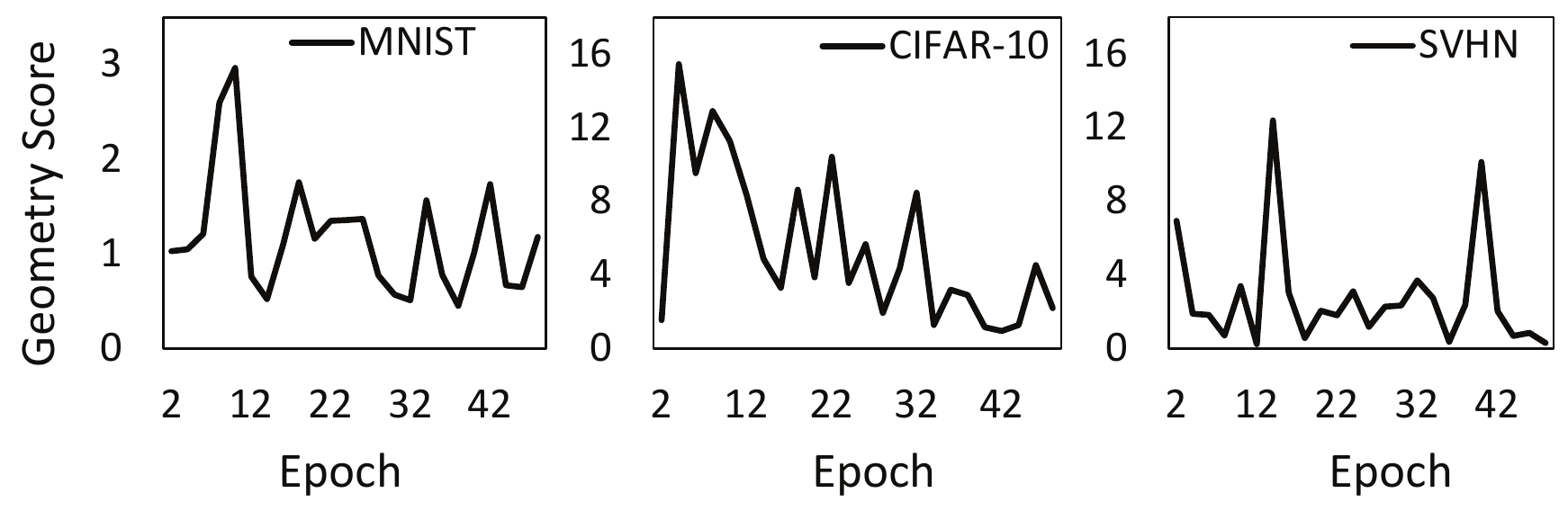} \\
\end{tabular}
\vspace{-0.1in}
\caption{Evaluation of Geometry score on sample quality for DCGAN models trained on MNIST (left), CIFAR-10 (middle) and SVHN (right). After each epoch of training (for the first 50 epochs), we computed the Geometry Score over 2000 samples generated by the network.}
\label{fig:geometryscore}
\end{figure}

\section{Sample Quality Evaluation of Geometry Score}
\label{sec:geometryscore} 
We train DCGANs on MNIST, CIFAR-10 and SVHN datasets, and compute the Geometry score using 2000 generated samples after each epoch of training for the first 50 epochs. The DCGAN architectures used here are the same as used in the other experiments of Section \ref{sec:experiments_crosslid} of the paper, and are described in Suppl. \ref{sec:expsetup}. As demonstrated in Figure~\ref{fig:geometryscore}, the Geometry Score exhibits high variation and has no clear correlation with the training progress and sample quality, which is consistent with the claims in \cite{gscore}. Visual verification of the improvement in sample quality over epochs can be found in Figures~\ref{fig:cifar10samplelidcorrelation} and \ref{fig:mnistsamplelidcorrelation}. Although a larger sample size (10000) was used in the original paper, we found it is computationally expensive to compute Geometry scores with such a large sample size over many epochs.

\section{Benefits of CrossLID Estimation in Deep Feature Space}
\label{sec:benefitsfslid}
\begin{figure}[ht!]
\centering
\setlength\tabcolsep{5.0pt}
\begin{tabular}{ccc}
   \includegraphics[width=35mm]{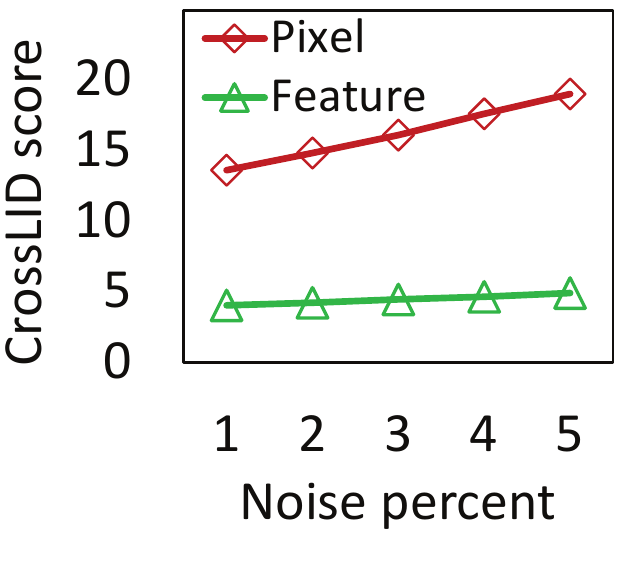} &
    	\includegraphics[width=35mm]{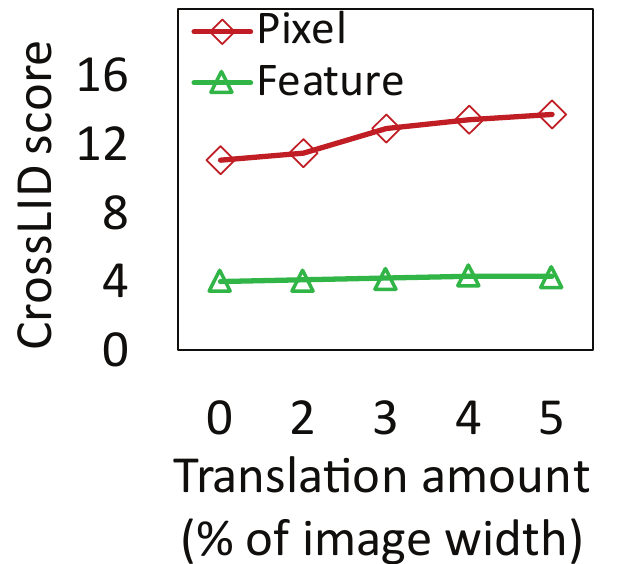} &
            \includegraphics[width=35mm]{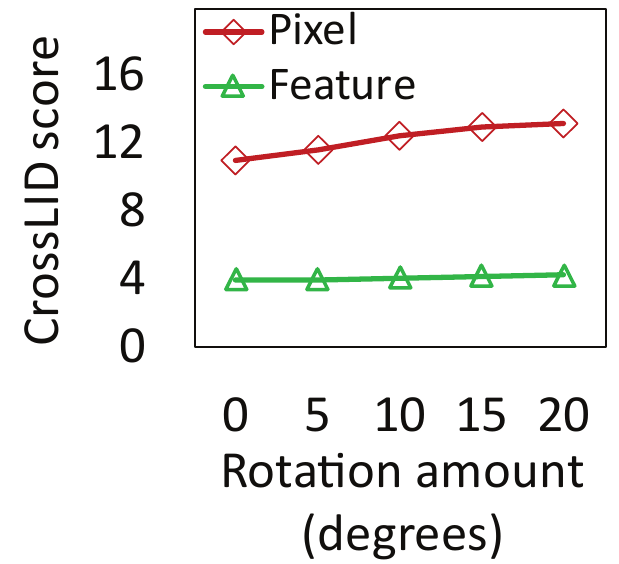} \\
\end{tabular}
\vspace{-0.1in}
\caption{Comparison of CrossLID scores estimated in the pixel space versus those estimated in the deep feature space, on the MNIST dataset, against small-scale salt-and-pepper (impulse) noise (left) and image transformations including translations (middle) and rotations (right).}
\label{fig:lidstudypsvsfsfliptrans}
\end{figure}

In Section \ref{sec:cross_estimation} of the paper, we proposed the use of deep feature space for the estimation of CrossLID. Following the discussion in Section \ref{sec:experiments_crosslid}, we show here experimental details justifying the advantages of CrossLID when computed within deep feature spaces. We test the estimation of CrossLID on purely real samples (that is, $\text{CrossLID}(X_{I}, X_{I})$) on the MNIST dataset for two settings: 1) directly in the pixel space, or 2) in deep feature space as defined by an external CNN classifier. Figure~\ref{fig:lidstudypsvsfsfliptrans} demonstrates the robustness of CrossLID in the two settings for three scenarios: 1) small scale input noise, 2) translations, and 3) rotations. The CrossLID scores estimated in the feature space exhibits less variation than those computed in the pixel space. Such robustness is expected from an evaluation metric which should not degrade highly with respect to small perturbations in the image pixels. This confirms the denoising and representation learning capabilities of convolution networks, and the advantage of CrossLID estimation in the deep feature space.

\section{Correlation between CrossLID and FID}
\label{sec:crosslidfidcorr}
\begin{figure}[h!]
\centering
\begin{tabular}{cc}
	\includegraphics[width=38mm]{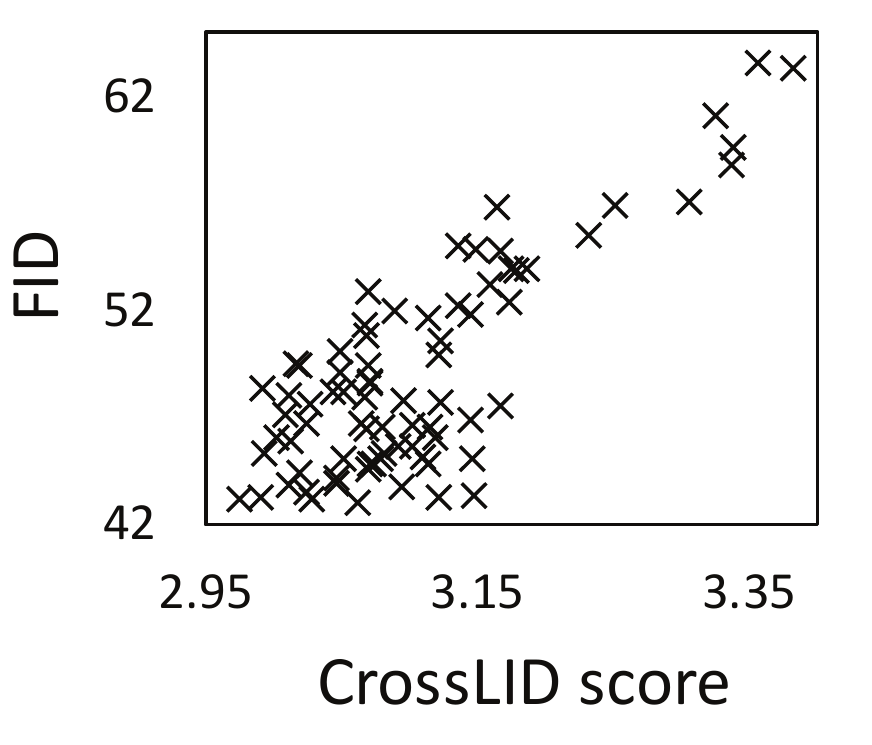} &
    	\includegraphics[width=36mm]{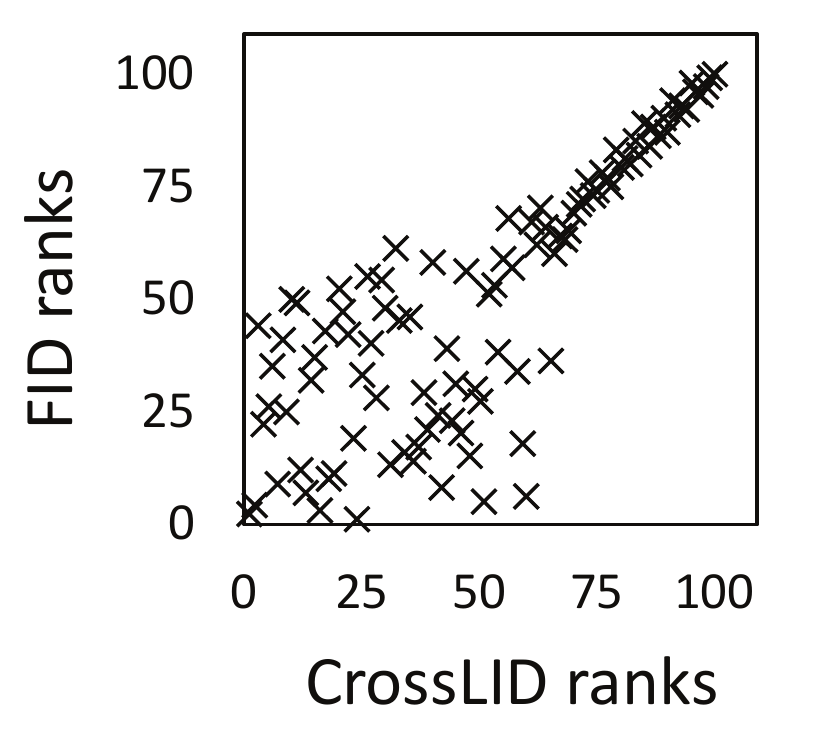}
\end{tabular}
\vspace{-0.1in}
\caption{\emph{Left}: Scatter plot of FID and CrossLID scores of different models over the CIFAR10 dataset. \emph{Right}: Scatter plot of FID and CrossLID ranks of different models over the CIFAR10 dataset.}
\label{fig:clidfidscatterplotcifar10}
\end{figure}

In Sec.~\ref{sec:experiments_crosslid} of the main paper, we have demonstrated that CrossLID exhibits a strong correlation with IS metric. Here, we evaluate the correlation between the CrossLID and FID metrics. We calculate both metrics over different models of the MNIST, CIFAR10, and SVHN datasets. The models were taken from different epochs of a DCGAN training on the datasets. For each dataset, we compute two correlations between the FID and CrossLID scores (of different epochs):  Pearson's correlation coefficient and Spearman's rank correlation coefficient. Table~\ref{tab:correlationclidfid} shows the computed value of the coefficients. We observe a strong (but not perfect) correlation between the two metrics across all three datasets in terms of both correlation measures. 
Figure~\ref{fig:clidfidscatterplotcifar10} shows the scatter plots of the scores for CIFAR10 dataset.   Based on these results, we can say that the two measures show considerable level of agreement, but differences in the assessment of quality are nevertheless present, particularly for the more complex dataset CIFAR10 (correlations of roughly 0.8).

\begin{table}[h!]
	\small
    \setlength\tabcolsep{5.0pt}
    \begin{center}
    \caption{Correlation coefficient between FID and CrossLID scores over different models across three datasets.}
    \vspace{0.1in}
    \label{tab:correlationclidfid}
    \begin{tabular}{c|c|c|c}
        \hline 
        &MNIST&CIFAR10&SVHN\\ \hline
        Pearson&0.99&0.79&0.69\\ \hline
        Spearman rank&0.98&0.81&0.97\\ \hline
    \end{tabular}
    \end{center}
\end{table}

\section{Visual Inspections on CrossLID score and Sample Quality}
\label{sec:correlationlidsamplequality}
\begin{figure*}[ht!]
\centering
\begin{tabular}{ccc}
	\includegraphics[width=35mm]{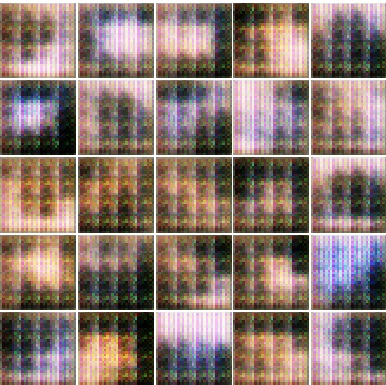} &
    	\includegraphics[width=35mm]{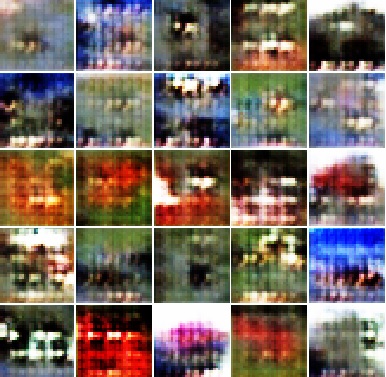} &
        	\includegraphics[width=35mm]{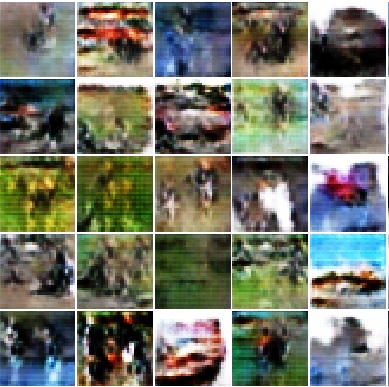} \\
    (a) CrossLID score = 30.8 & (b) CrossLID score = 7.24 & (c) CrossLID score = 4.18 \\
    \includegraphics[width=35mm]{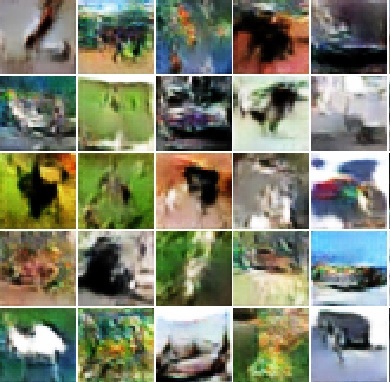} &
        \includegraphics[width=35mm]{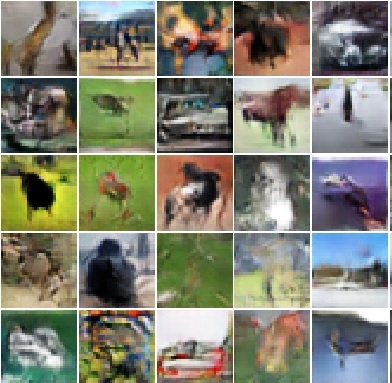} & 
            \includegraphics[width=35mm]{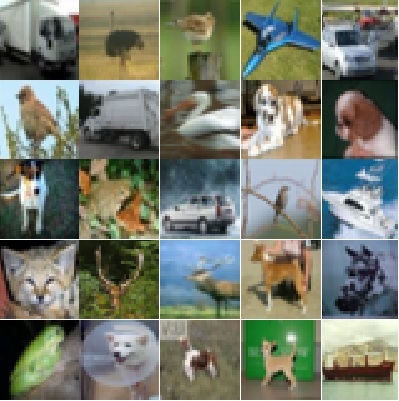} \\
    (d) CrossLID score = 3.76 & (e) CrossLID score = 3.04 & (f) CrossLID score = 1.34 \\        
\end{tabular}
\caption{(a-e) 25 randomly selected DCGAN generated CIFAR-10 images and the CrossLID score of the DCGAN model after epoch 1, 5, 10, 20, and 49; (f) Real CIFAR-10 images and the $\text{CrossLID}(X_I; X_I)$ score.}
\label{fig:cifar10samplelidcorrelation}
\end{figure*}

\begin{figure*}[ht!]
\centering
\begin{tabular}{ccc}
	\includegraphics[width=30mm]{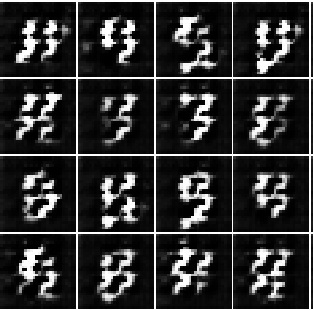} &
    	\includegraphics[width=30mm]{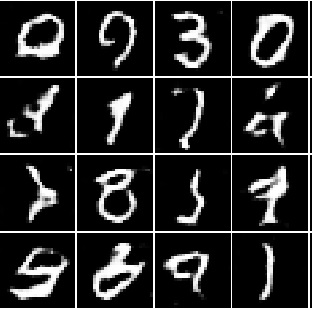} &
        	\includegraphics[width=30mm]{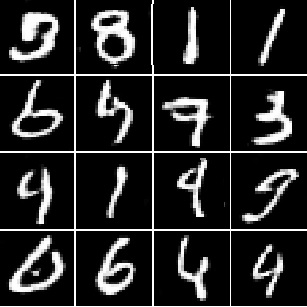} \\
    (a) CrossLID score = 28.3 & (b) CrossLID score = 7.12 & (c) CrossLID score = 5.88 \\
    \includegraphics[width=30mm]{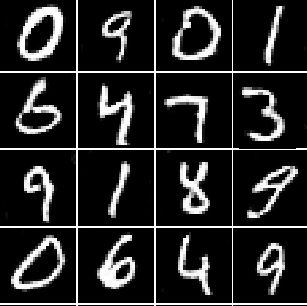} &
        \includegraphics[width=30mm]{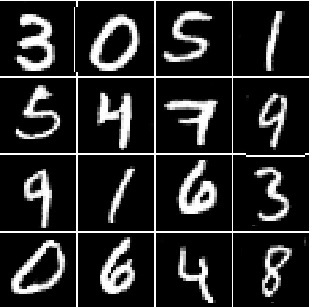} & 
            \includegraphics[width=30mm]{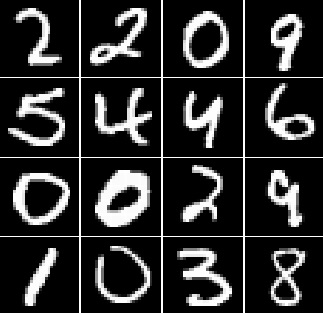} \\
    (d) CrossLID score = 5.36 & (e) CrossLID score = 5.06 & (f) CrossLID score = 3.90 \\       
\end{tabular}
\caption{(a-e) 16 randomly selected DCGAN generated MNIST images and the CrossLID score of the DCGAN model after epoch 1, 4, 10, 20, and 50; (f) Real MNIST images and the $\text{CrossLID}(X_I; X_I)$ score.}
\label{fig:mnistsamplelidcorrelation}
\end{figure*}

In Section \ref{sec:experiments_crosslid} of the paper, we demonstrated that the CrossLID score is closely correlated with the progress of GAN training, in that the CrossLID score decreases consistently as the model progressively learns to generate samples of better quality. Here, we visually inspect some sample images generated at different training stages of a DCGAN model, on the MNIST and CIFAR-10 datasets. Figures~\ref{fig:cifar10samplelidcorrelation} and \ref{fig:mnistsamplelidcorrelation} show examples of generated CIFAR-10 and MNIST images respectively. As can be seen, the sample quality increases as training proceeds, and there is a strong correlation between decreasing CrossLID score and increasing sample quality. Note that the last subfigures in both Figure~\ref{fig:cifar10samplelidcorrelation} and Figure~\ref{fig:mnistsamplelidcorrelation} show the real images and the CrossLID score of the real data distribution ($\text{CrossLID}(X_I; X_I)$), which is the lowest overall.

\begin{figure}[ht!]
\centering
\begin{tabular}{c}
    	\includegraphics[width=0.65\textwidth]{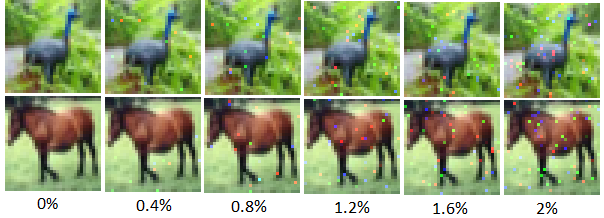} \\		    \includegraphics[width=0.35\textwidth]{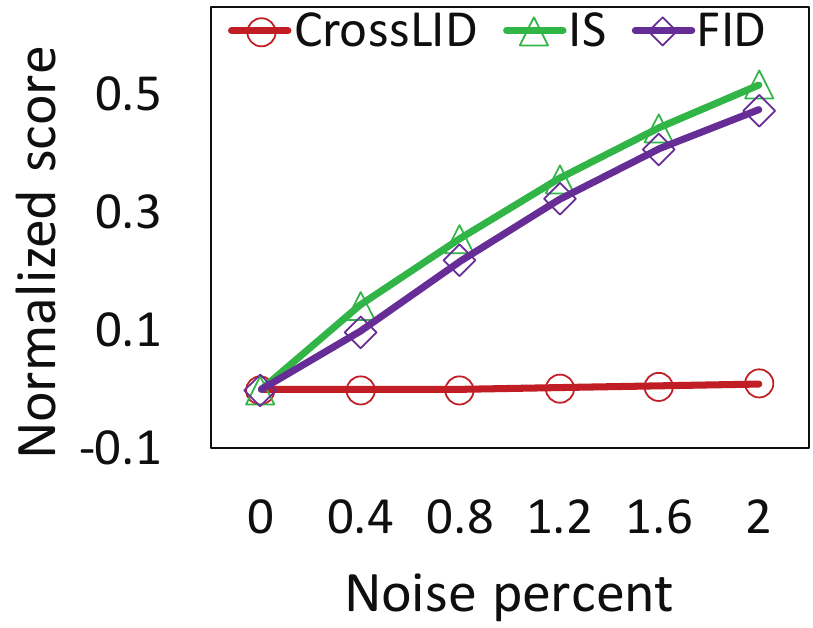} \\ 
\end{tabular}
\caption{ \emph{Top}: Some representative CIFAR10 images after the application of different percentage of Gaussian noise. \emph{Bottom}: Normalized CrossLID score, IS, and FID under different levels of Gaussian noise.}
\label{fig:scorenormalizedGaussiancifar}
\end{figure}

\begin{figure}[ht!]
\centering
\begin{tabular}{cc}
    	\includegraphics[width=0.35\textwidth]{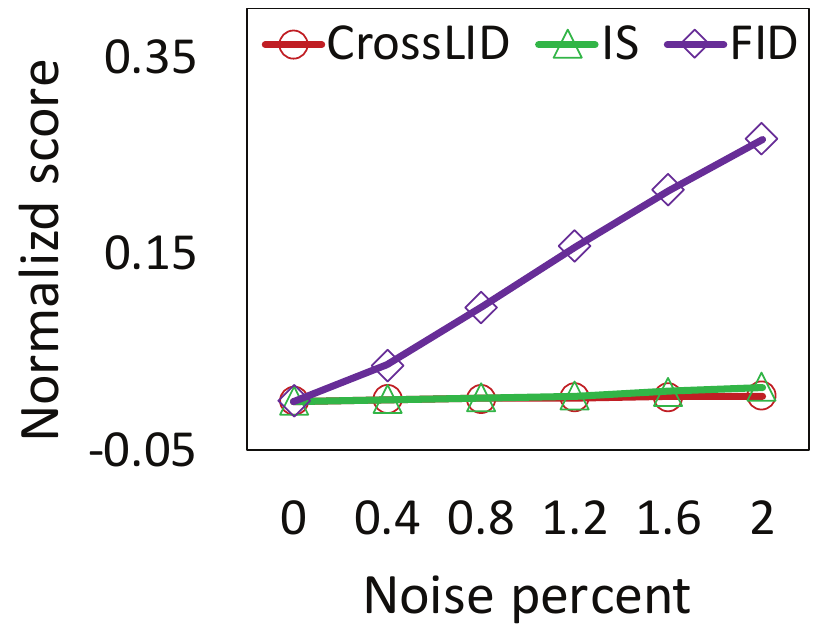} &
    	        \includegraphics[width=0.33\textwidth]{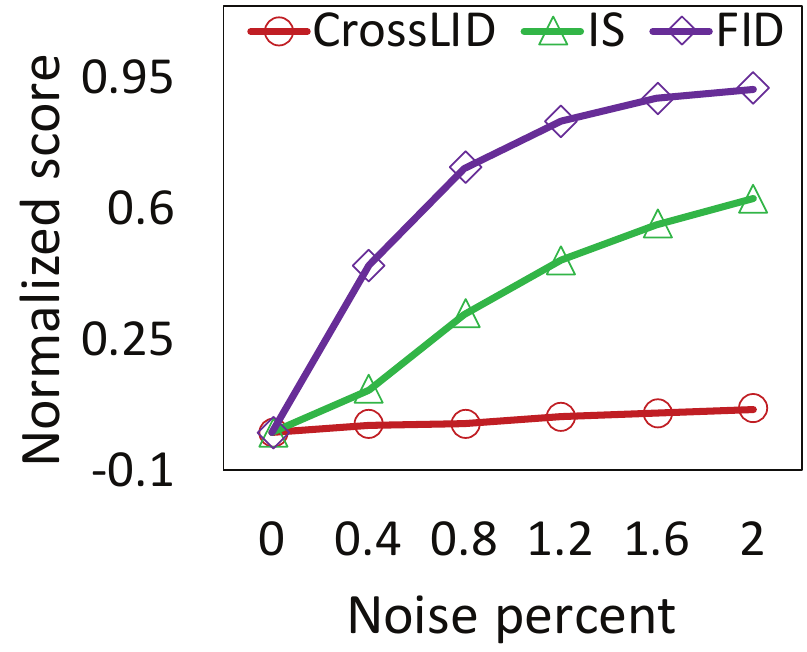} \\
\end{tabular}
\caption{Normalized CrossLID score, IS, and FID under different levels of Gaussian noise on MNIST (\emph{Left}) and SVHN (\emph{Right}) datasets.}
\label{fig:scorenormalizedGaussian}
\end{figure}

\section{Comparison of CrossLID, IS, and FID using Normalized Scores}
\label{normalizedscoresnoise}
In Sec. \ref{sec:experiments_crosslid}, we compared the robustness of CrossLID, IS, and FID metrics against small Gaussian noise. We compared the raw score values of the three metrics for different levels of input noise (small) and found CrossLID to be more robust than the other two metrics. Here, we provide further insights using a relative comparison of the three metrics in terms of their normalized score values. We normalize the raw scores of all three metrics to $[0,1]$ by $\frac{S-S_0}{S_{100}-S_0}$ where $S$ is the score at a specific noise level, $S_0$ is the score at no noise, and $S_{100}$ is the score at maximum (100\%) noise. Note that for all metrics, a lower normalized score indicates better performance. 

The top subfigure of Figure~\ref{fig:scorenormalizedGaussiancifar} shows representative CIFAR10 images at various noise levels, confirming that the visual quality of the images is affected only slightly through the addition of noise.
The bottom subfigure of Figure~\ref{fig:scorenormalizedGaussiancifar} shows the normalized scores for different levels of Gaussian noise (up to 2\%). We find that both IS and FID have disproportionately high sensitivity to low level noise. For example, IS and FID scores changes by 52\% and 48\%, respectively, at 2\% noise. In contrast, the response of CrossLID is much more proportionate, a change of 1.2\% (at 2\% noise level). Moreover, considering that the images suffers almost no distortion and are highly recognizable, the response of CrossLID seems more reasonable than the response of the other two metrics.
Similar results for MNIST and SVHN datasets can be found in Figure~\ref{fig:scorenormalizedGaussian}. 
In our opinion, a high sensitivity to this low level of noise is quite undesirable, since the noisy images are highly recognizable. We note however that such a conclusion might be regarded subjective, as the GAN research community has not yet come to a consensus as to whether it is desirable for a quality measure to have high sensitivity to a level of noise which does not substantially alter the visual quality of an image.  As a further comparison, in Sec. \ref{sec:robustnessspnoise} we study the behavior of GAN quality measures in the presence of very high levels of input noise.

\section{Further Robustness Evaluation on Input Noise}
\label{sec:robustnessspnoise} 
\begin{figure}[ht!]
\centering
\begin{tabular}{c}
    	\includegraphics[width=0.65\textwidth]{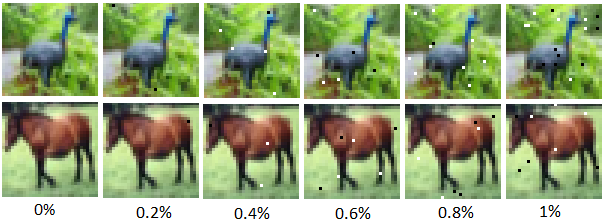} \\
    	    \includegraphics[width=0.35\textwidth]{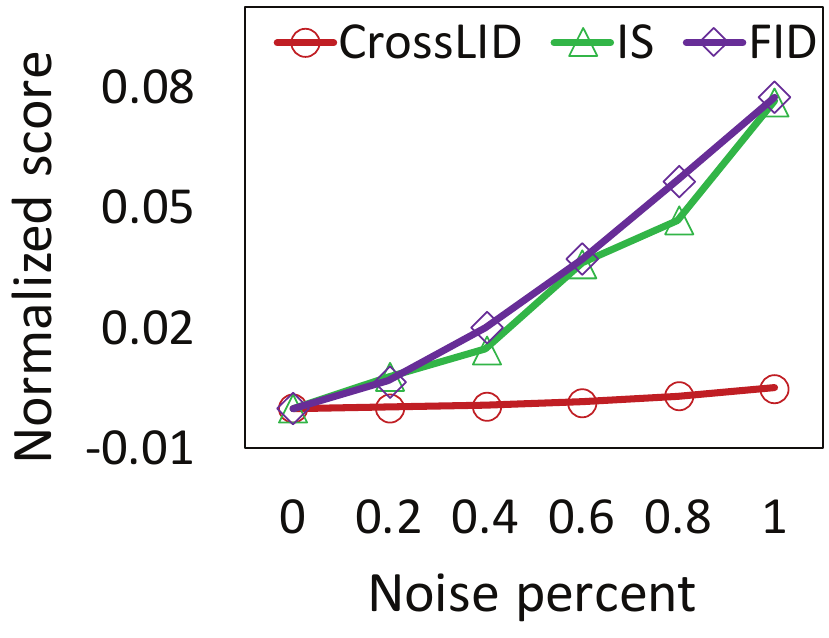} \\ 
\end{tabular}
\caption{ \emph{Top}: representative CIFAR10 images after the application of different percentages of salt-and-pepper noise. \emph{Bottom}: Normalized CrossLID score, IS, and FID  under low levels of salt-and-pepper noise.}
\label{fig:lidinceptionfidvaryspnoise}
\end{figure}

In addition to the evaluation of robustness against Gaussian input noise presented in Figure~\ref{fig:lidinceptionfidvarytrans} of Section \ref{sec:experiments_crosslid} of the main text, we provide here some analysis on one form of real-world image noise, the so-called `salt-and-pepper' (or impulse) noise. The top subfigure of Figure~\ref{fig:lidinceptionfidvaryspnoise} shows some representative images for low level salt-and-pepper noise, and the bottom subfigure shows the corresponding normalized scores of the three metrics CrossLID, IS, and FID. 
As shown in the top subfigure of Figure~\ref{fig:lidinceptionfidvaryspnoise}, the images are only slightly affected by the low levels of noise applied in the tests. For these low levels of noise, both the FID and IS scores increase significantly, whereas CrossLID exhibits small and proportionate variation. 

\begin{figure}[ht!]
\centering
\begin{tabular}{cc}
    \includegraphics[width=0.28\textwidth]{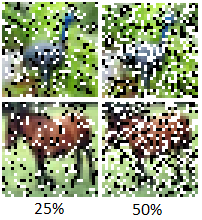} &
	    \includegraphics[width=0.35\textwidth]{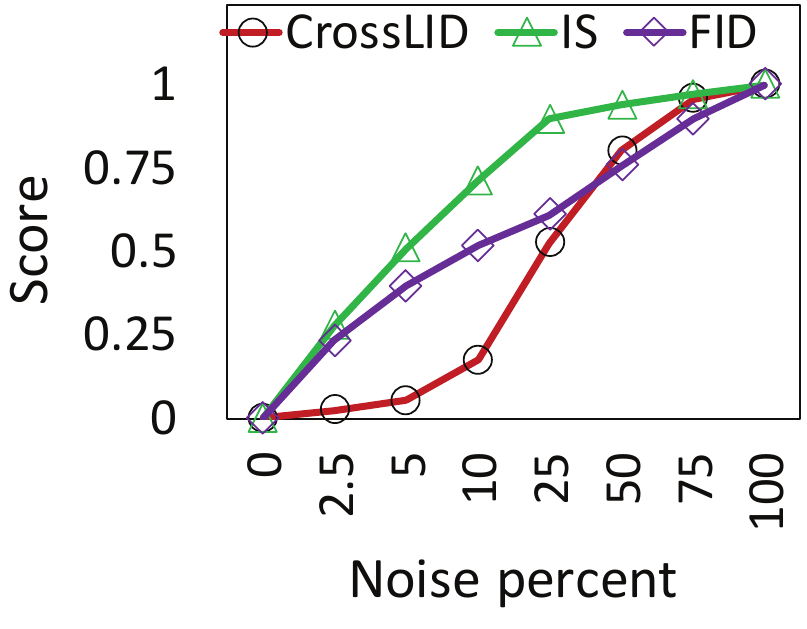} \\
\end{tabular}
\caption{\emph{Left}: Some representative CIFAR10 images with 25\% and 50\% salt-and-pepper noise. \emph{Right}: Normalized CrossLID score, IS, and FID for images with high levels of salt-and-pepper noise.  Note the non-linear scale on the $x$-axis.}
\label{fig:lidinceptionfidvarylargespnoise}
\vspace{-0.1in}
\end{figure}

We further show, in Figure~\ref{fig:lidinceptionfidvarylargespnoise}, how the three metrics respond to the entire range of salt-and-pepper noise from 0\% to 100\%. Although all three scores increase as the noise level increases to 100\%, they react in quite different ways. Both IS and FID are highly sensitive to low noise levels, while CrossLID is less sensitive. CrossLID increases sharply as noise begins to dominate (as seen in our results for 25\% and 50\% noise); from the representative images on the left, we see that this effect coincides with a drastic drop in visual quality. We believe that the response of CrossLID to different noise levels is more reasonably correlated with the visual quality of images, but note that this evaluation is necessarily subjective.

\section{Comparison of Running Time}
\label{sec:metricexectime}
\begin{figure}[ht!]
\centering
\begin{tabular}{cc}
	\includegraphics[width=0.35\textwidth]{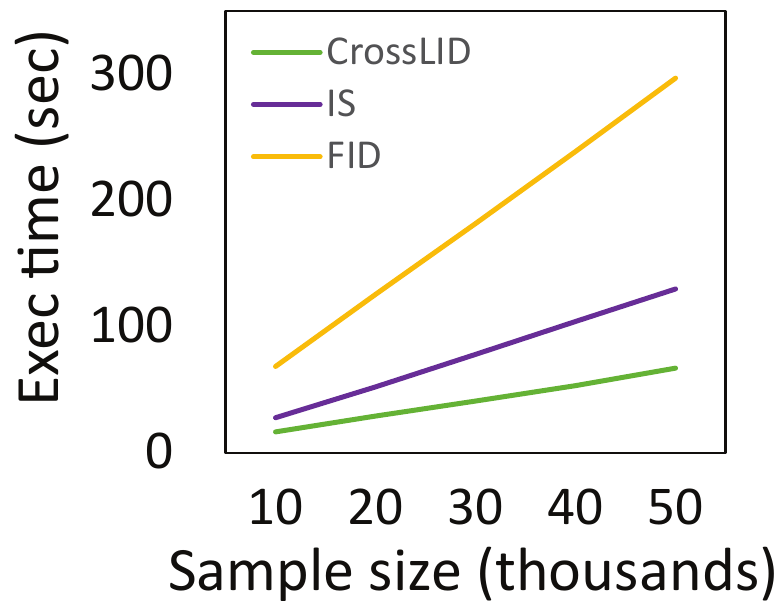} &
	\includegraphics[width=0.35\textwidth]{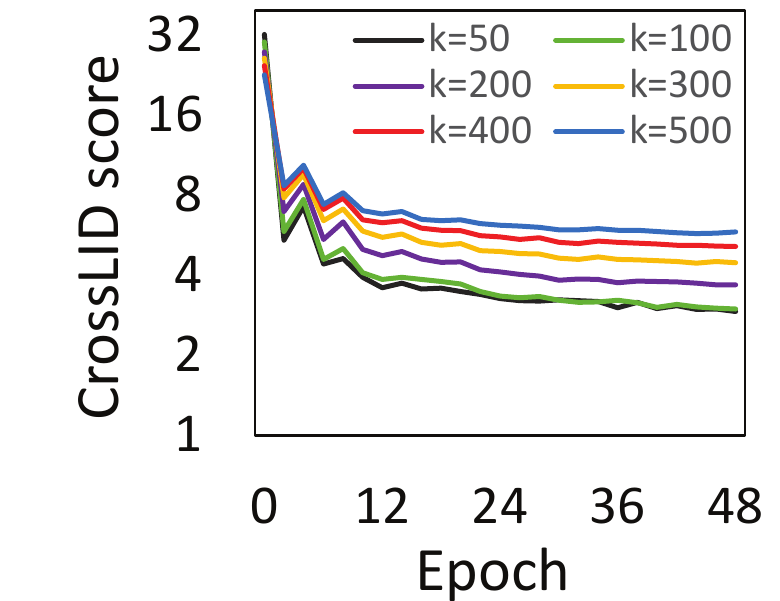} \\
\end{tabular}
\caption{\emph{Left}: Running time of CrossLID, IS and FID over different sample sizes on the CIFAR-10 dataset. \emph{Right}: CrossLID scores estimated using different neighborhood size $k$, for images generated at different epochs by a DCGAN on the CIFAR-10 dataset.}
\label{fig:metricexecutiontimeclidvaryk}
\vspace{-0.1in}
\end{figure}

We compare the running time of CrossLID, IS, and FID with respect to different sample sizes (from 10K to 50K) used for estimation. The running time was calculated as the time elapsed to obtain a score for a given set of samples, with an NVIDIA Titan V GPU. The left subfigure of Figure~\ref{fig:metricexecutiontimeclidvaryk} shows the result on the CIFAR-10 dataset. As the sample size increases, the running time of the three metrics all increase in a linear fashion, but at different pace and scales. Among them, the CrossLID metric has the lowest running time while FID requires the highest. 
Note that the different feature extractors used by the three metrics may influence their running time slightly.

\section{Impact of Neighborhood Size on CrossLID Estimation}
\label{sec:crosslidvaryk}
CrossLID has an additional hyper-parameter $k$ which is the size of the local neighborhood used for estimation. 
For all experiments and datasets used in our experiments, we found $k=100$ to be a good neighborhood size. The impact of varying $k$ on CrossLID estimation is further illustrated in the right subfigure of Figure \ref{fig:metricexecutiontimeclidvaryk}. For different choices of $k$, the estimated CrossLID scores all decrease as training progresses, which indicates that the estimates all accurately captured the improvement of GAN models over training. We also observe that a larger $k$ tends to result in a higher value of the estimate, an effect of the expansion of locality. Overall, the discriminability of CrossLID is not greatly sensitive to variation in the value of $k$..

\begin{table}[h!]
    \small
    \setlength\tabcolsep{5.0pt}
    \begin{center}
    \caption{Inception scores (IS) (mean$\pm$std over 5 random runs) of CrossLID guided oversampling.}
    \vspace{0.1in}
    \label{tab:inceptioncrosslidexp}
    \begin{tabular}{c|c|c||c|c}
        \hline
      	&\multicolumn{4}{c}{IS (higher is better)} \\ \hline
        Dataset&DCGAN&DCGAN+&WGAN&WGAN+ \\ \hline
        MNIST&8.65 $\pm$ 0.03&\textbf{8.76 $\pm$ 0.01}&8.13 $\pm$ 0.04&\textbf{8.46 $\pm$ 0.03}\\ \hline
        CIFAR10&6.14 $\pm$ 0.09&\textbf{6.32 $\pm$ 0.09}&5.46 $\pm$ 0.03&\textbf{5.70 $\pm$ 0.06}\\ \hline
        SVHN&\textbf{3.03 $\pm$ 0.03}&3.01 $\pm$ 0.02&\textbf{2.90 $\pm$ 0.02}&2.89 $\pm$ 0.01 \\ \hline
    \end{tabular}
    \end{center}
    \vspace{-0.1in}
\end{table}

\begin{table}[h!]
    \small
    \setlength\tabcolsep{5.0pt}
    \begin{center}
    \caption{FID scores (mean$\pm$std over 5 random runs) of CrossLID guided oversampling.}
    \vspace{0.1in}
    \label{tab:fidcrosslidexp}
    \begin{tabular}{c|c|c||c|c}
        \hline
      	&\multicolumn{4}{c}{FID (lower is better)} \\ \hline
        Dataset&DCGAN&DCGAN+&WGAN&WGAN+ \\ \hline
        MNIST&7.13 $\pm$ 0.02&\textbf{6.49 $\pm$ 0.04}&16.40 $\pm$ 0.06&\textbf{12.44 $\pm$ 0.17}\\ \hline
        CIFAR10&40.98 $\pm$ 0.24&\textbf{39.75 $\pm$ 0.12}&56.06 $\pm$ 0.24&\textbf{54.26 $\pm$ 0.22}\\ \hline
        SVHN&13.87 $\pm$ 0.07&\textbf{12.92 $\pm$ 0.07}&35.86 $\pm$ 0.12&\textbf{32.82 $\pm$ 0.20}\\ \hline
    \end{tabular}
    \end{center}
    \vspace{-0.1in}
\end{table}

\section{Evaluation of CrossLID-Guided Oversampling Approach using IS and FID}
\label{sec:inceptionfidoversamplingexp}
Table~\ref{tab:inceptioncrosslidexp} and~\ref{tab:fidcrosslidexp} report the results in terms of IS and FID, of the CrossLID guided oversampling experiments described in Section~\ref{sec:agumented_training}. We see that CrossLID guided oversampling methods (DCGAN+ and WGAN+) achieve lower FID results than those of standard GAN training (DCGAN and WGAN). Comparing the FID results with the CrossLID scores reported in Table~\ref{tab:dcganlidinceptionscores}, we find that their rankings are the same: the proposed oversampling approach achieved better results in terms of both metrics. This is the case with IS as well, except for the SVHN dataset, for which both methods demonstrate similar results.

\begin{figure}[ht!]
	\centering
	\small
	\begin{tabular}{ccc}
		\includegraphics[width=22mm]{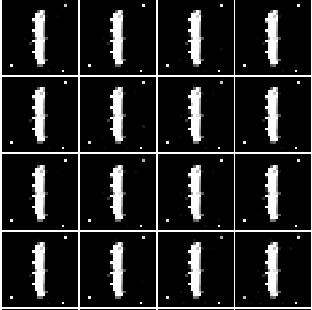} & 
			\includegraphics[width=22mm]{img/stabilitystudy/nbnnbn/stdgan/epoch30.jpg} & 
				\includegraphics[width=22mm]{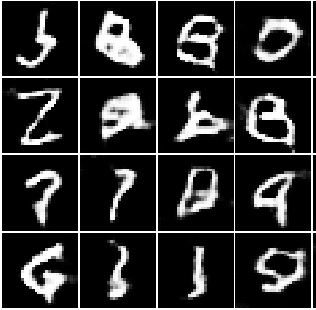} \\ 
		(a) Epoch 20 & (b) Epoch 30 & (c) Epoch 50 \\
		\includegraphics[width=22mm]{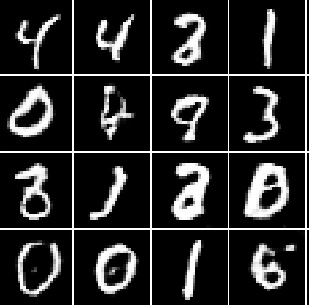} & 
			\includegraphics[width=22mm]{img/stabilitystudy/nbnnbn/lidoptgan/epoch30.jpg} & 
				\includegraphics[width=22mm]{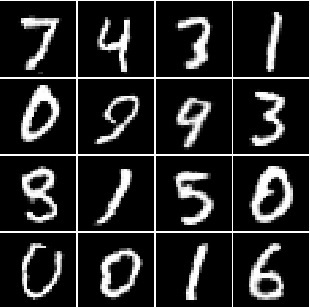} \\ 
		(d) Epoch 20 & (e) Epoch 30 & (f) Epoch 50 \\
\end{tabular}
\caption{MNIST images generated at epoch 20, 30 and 50 (50 epochs in total) by DCGAN models without batch normalization layers in both the generator and discriminator. \emph{Top row}: Images generated by a DCGAN model trained using standard training. \emph{Bottom row}: Images generated by a DCGAN model trained with our proposed oversampling strategy.}
\label{fig:nbnnbnmnistoutputs}
\end{figure}

\begin{figure}[ht!]
	\centering
	\small
	\begin{tabular}{ccc}
		\includegraphics[width=22mm]{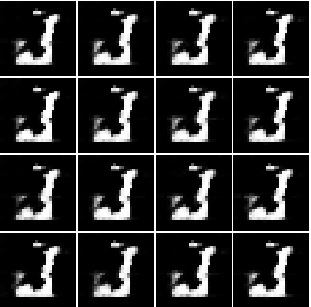} & 
			\includegraphics[width=22mm]{img/stabilitystudy/bnnbn/stdgan/epoch30.jpg} & 
				\includegraphics[width=22mm]{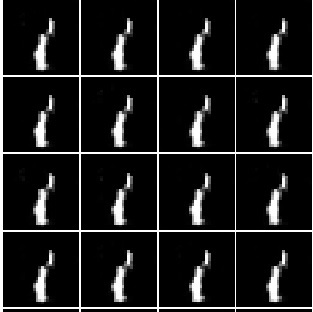} \\ 
		(a) Epoch 20 & (b) Epoch 30 & (c) Epoch 50 \\
		\includegraphics[width=22mm]{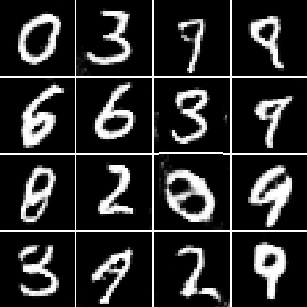} & 
			\includegraphics[width=22mm]{img/stabilitystudy/bnnbn/lidoptgan/epoch30.jpg} & 
				\includegraphics[width=22mm]{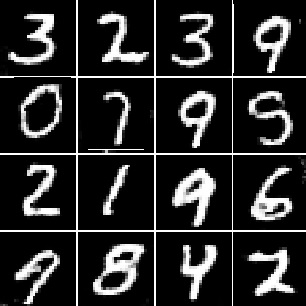} \\ 
		(d) Epoch 20 & (e) Epoch 30 & (f) Epoch 50 \\
\end{tabular}
\vspace{-0.1in}
\caption{MNIST images generated at epoch 20, 30 and 50 (50 epochs in total) by DCGAN models without batch normalization layers in the discriminator (the generator network still has batch normalization). \emph{Top row}: Images generated by a DCGAN model trained using standard training. \emph{Bottom row}: Images generated by a DCGAN model trained with our proposed oversampling strategy.}
\label{fig:bnnbnmnistoutputs}
\vspace{-0.1in}
\end{figure}

\section{Stability of the Proposed GAN Training with Oversampling}
\label{sec:stability}
Here, we provide more details regarding the training of DCGANs without batch normalization layers 1) in the discriminator, or 2) in both the generator and the discriminator. Following the methodology used in \cite{wgan} to verify model stability, we illustrate how our proposed oversampling strategy can avoid mode collapse and help learning, by showing for the MNIST dataset the images generated at different epochs. As visualized in Figure~\ref{fig:nbnnbnmnistoutputs}, when batch normalization layers were removed from both the generator and the discriminator, standard training suffered from mode collapse from the beginning of training, and the images generated images by the end of training were of low quality. More severe mode collapse was observed with standard training when the batch normalization layers were removed from the discriminator only: in this case, the model failed to generate realistic images (see Figure~\ref{fig:bnnbnmnistoutputs}). On the other hand, when trained with our proposed oversampling strategy, mode collapse was not observed in any of the two scenarios, and the generated images were of higher quality consistently throughout training.


\section{Network Architectures and Experimental Settings}
\subsection{GAN architecture, Training Details and Exemplary Outputs}
\label{sec:expsetup}
For all experiments, we constructed GAN models based on the architectural guidelines of DCGAN~\cite{dcgan}. The generator and discriminator networks used for MNIST and CIFAR-10/SVHN are described in Table \ref{tab:cnnmodelmnist} and Table \ref{tab:cnnmodelcifar10} respectively. The same network architecture was used for WGAN experiments except that the output activation (sigmoid) in the discriminator was removed. This was done to produce a linear output for the training of WGANs with the Wasserstein loss \cite{wgan}.

\begin{figure}[ht!]
    \setlength\tabcolsep{3pt}
	\centering
	\begin{tabular}{ccc}
		\includegraphics[width=25mm]{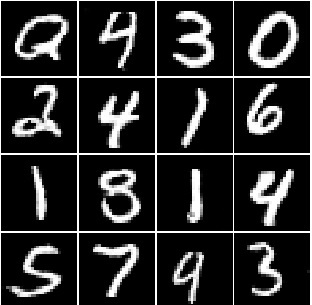} & 
		    \includegraphics[width=25mm]{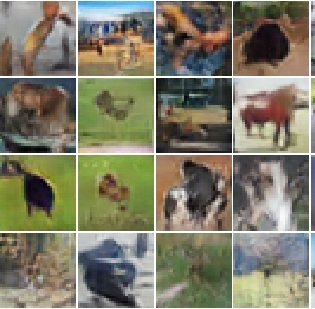} &
		    	\includegraphics[width=25mm]{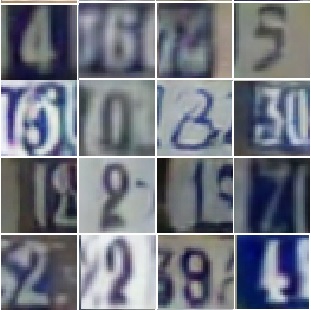} \\
		(a) MNIST & (b) CIFAR-10 & (c) SVHN  \\

		\includegraphics[width=25mm]{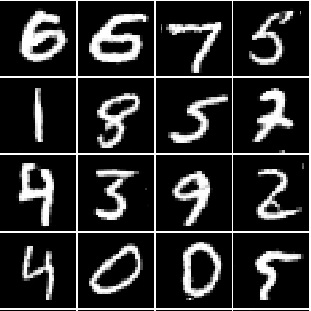} &  
			\includegraphics[width=25mm]{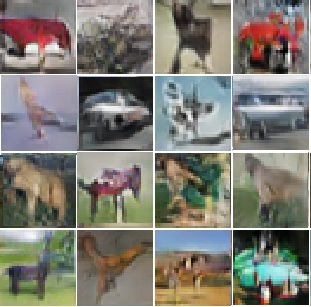} &
				\includegraphics[width=25mm]{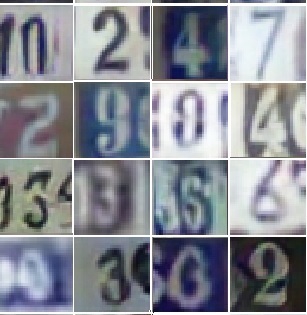}  \\ 
		(d) MNIST & (e) CIFAR-10 & (f) SVHN \\
\end{tabular}
\caption{\emph{Top row:} Images generated by standard DCGAN training (without our proposed oversampling)  on the three datasets (a-c). \emph{Bottom row:} Images generated by DCGANs trained with our proposed oversampling strategy on the three datasets (d-f).}
\label{fig:ganoutputs}
\end{figure}

For DCGAN training, the models were trained for 60 and 100 epochs on MNIST and CIFAR-10/SVHN respectively, using the Adam optimizer \cite{adam} with learning rate 0.0001 and decay 0.00001. The WGANs models were trained for 200 epochs on all datasets using the RMSProp optimizer.
A learning rate of 0.00005 was used for CIFAR10 and SVHN training, while 0.0001 was used for MNIST. 
For our proposed CrossLID guided training approach, we used $N_1=20K, N_2=2K$ for class-wise CrossLID estimation, and sample size $m=30K$ for DCGANs and $m=20K$ for WGANs. For $T$, we simply used the number of generator iterations in one epoch, i.e., we applied oversampling after every epoch. Figure~\ref{fig:ganoutputs} shows some randomly selected images generated by 1) DCGANs trained using standard training versus 2) DCGANs trained with our proposed oversampling strategy.

\begin{table}[ht!]
    \small
    \renewcommand{\arraystretch}{1.1}
    \begin{center}
    \caption{The generator and discriminator network used for the MNIST dataset. Conv($x,y,z$) represents a convolution layer with $x$ filters of kernel size $y \times y$ and stride $z$. ConvTr($x,y,z$) represents a transposed convolution layer with $x$ filters of kernel size $y \times y$ and stride $z$. FC($x$) represents a fully connected layer with $x$ output nodes. BN represents a batch normalization layer, R represents the reshape operation and LReLU is the LeakyRelu.}
    \vspace{0.1in}
    \label{tab:cnnmodelmnist}
    \begin{tabular}{c|c}
        \multicolumn{2}{c}{} \\ \hline 
        Generator&Discriminator\\ \hline
        Input: Z(100)&Input: (28,28,1)\\
        R(1,1,100)&Conv(32,3,2), BN, LReLU\\
        ConvTr(128,3,1), BN, ReLU&Conv(64,3,2), BN, LReLU\\
        ConvTr(64,3,2), BN, ReLU&Conv(128,3,2), BN, LReLU\\
        ConvTr(32,3,2), BN, ReLU&Conv(1,3,1), Sigmoid\\
        ConvTr(1,3,2), Tanh&Output: 1\\
        Output: (28, 28, 1)&\\ \hline
    \end{tabular}
  \end{center}
  \vspace{-0.1in}
\end{table}

\begin{table}[ht!]
    \small
    \renewcommand{\arraystretch}{1.1}
     \begin{center}
    \caption{The generator and discriminator network used for the CIFAR-10 and SVHN datasets. Conv($x,y,z$) represents a convolution layer with $x$ filters of kernel size $y \times y$ and stride $z$. ConvTr($x,y,z$) represents a transposed convolution layer with $x$ filters of kernel size $y \times y$ and stride $z$. FC($x$) represents a fully connected layer with $x$ output nodes. BN represents a batch normalization layer, R represents the reshape operation and LReLU is the LeakyRelu}
    \vspace{0.1in}
    \label{tab:cnnmodelcifar10}
    \begin{tabular}{c|c}
        \multicolumn{2}{c}{} \\ \hline 
        Generator&Discriminator\\ \hline
        Input: Z(100)&Input: (32,32,3)\\
        R(1,1,100)&Conv(64,4,2), BN, LReLU\\
        ConvTr(256,4,1), BN, ReLU&Conv(128,4,2), BN, LReLU\\
        ConvTr(128,4,2), BN, ReLU&Conv(256,4,2), BN, LReLU\\
        ConvTr(64,4,2), BN, ReLU&Conv(1,4,1), Sigmoid\\
        ConvTr(3,4,3), Tanh&Output: 1\\
        Output: (32, 32, 3)&\\ \hline
    \end{tabular}
  \end{center}
  \vspace{-0.1in}
\end{table}

For evaluating the CrossLID scores on ImageNet dataset, ResNet architecture was used for building the generator and discriminator network of the GAN model. The model was trained using WGAN-GP algorithm for 100K generator iterations using the parameter settings described in~\cite{improvedwgan}. Table~\ref{tab:resnetimagenet} shows the architecture used for the generator and discriminator networks. For details on ResNet block architecure, please see~\cite{improvedwgan}.

\begin{table}[ht!]
    \small
    \renewcommand{\arraystretch}{1.1}
     \begin{center}
    \caption{ResNet architecture used for the ImageNet training. ResBlock(U,$x$) denotes an upsampling ResNet block with $x$ number of filters while ResBlock(D,$x$) denotes a downsampling ResNet block.}
    \label{tab:resnetimagenet}
    \begin{tabular}{c|c}
        \multicolumn{2}{c}{} \\ \hline 
        Generator&Discriminator\\ \hline
        Input: Z(100) & Input: (128,128,3)\\
        FC($4\times4\times128$) & ResBlock(D,64)\\
        ResBlock(U,1024) & RestBlock(D,128) \\
        ResBlock(U,512) & ResBlock(D,256)\\
        ResBlock(U,256) & ResBlock(D,512)\\
        ResBlock(U,128) & ResBlock(D,1024)\\
        ResBlock(U,64) & GlobalAvgPool \\
        BN, ReLU & FC(1) \\
        Conv(3,3,1), Tanh & Output: 1 \\
        Output: (128,128,3) & \\
    \end{tabular}
  \end{center}
  \vspace{-0.1in}
\end{table}

\subsection{External CNN models used for Feature Extraction}
\label{sec:cnnfeatureextractor}

Table \ref{tab:cnnfeatureextractor} describes the architectures of external CNN models used for feature extraction on MNIST, CIFAR-10, SVHN, and ImageNet datasets; the selected feature layers are highlighted in \textbf{bold}. These CNN classifiers were trained separately on the original training sets of the three datasets. For MNIST and SVHN networks, the outputs of the first fully connected (FC) layer was used as features, while for CIFAR-10, the output of the last max pooling layer was used. For ImageNet, the last pooling layer (global average pooling having 2048 outputs) of a pretrained Inception-v3~\cite{inceptionv3} network was used for feature extraction. For the estimation of our proposed CrossLID, the networks were applied to extract the features for both real and fake images, and the extracted features were then used to compute the CrossLID scores.

\begin{table}[t]
\renewcommand{\arraystretch}{1.1}
\caption{Network architecture of external CNN models used for feature extraction. Conv($x,y,z$) represents a convolution layer with $x$ filters of kernel size $y \times y$ and stride=$z$. MaxPool($x,y$) represents a max-pooling layer with pool size $x \times y$. FC($x$) represents a fully connected layer with $x$ output nodes. The selected feature layers are highlighted in \textbf{bold}.}
\vspace{0.1in}
\centering
    \small
\begin{tabular}{m{15mm}|m{62mm}}
    \hline
    Dataset & Architecture of external CNN models \\ \hline
    MNIST &  Conv(32,3,1), Conv(64,3,1), MaxPool(2,2), \textbf{FC(128)}, FC(10) \\ \hline
    SVHN & Conv(32,3,1), Conv(32,3,1), MaxPool(2,2), Conv(64,3,1), Conv(64,3,1), MaxPool(2,2),
	Conv(128,3,1), Conv(128,3,1), MaxPool(2,2), \textbf{FC(512)}, FC(10) \\ \hline
	CIFAR-10 &
		Conv(64,3,1), Conv(64,3,1), MaxPool(2,2), Conv(128,3,1), Conv(128,3,1), MaxPool(2,2)
		Conv(256,3,1), Conv(256,3,1), Conv(256,3,1), MaxPool(2,2), Conv(512,3,1), Conv(512,3,1),
		Conv(512,3,1), MaxPool(2,2), Conv(512,3,1), Conv(512,3,1), Conv(512,3,1), \textbf{MaxPool(2,2)},
		FC(512), FC(10) \\ \hline
	ImageNet & Inception-v3 network, \textbf{the last pooling layer} (global average pooling having 2048 outputs) used for feature extraction \\ \hline
\end{tabular}
\label{tab:cnnfeatureextractor}
\vspace{0.1in}
\end{table}

\end{document}